\documentclass[11pt]{article}
\usepackage[margin=1in]{geometry}
\usepackage{amsmath,amssymb,amsthm}
\usepackage{graphicx}
\usepackage{booktabs}
\usepackage{multirow}
\usepackage[hidelinks]{hyperref}
\usepackage{xcolor}
\usepackage{natbib}
\let\cite\citep
\newtheorem{proposition}{Proposition}
\newtheorem{definition}{Definition}
\newtheorem{remark}{Remark}

\title{When to Ask for Help: A Decision-Theoretic Account and\\
Measurement Standard for Slow-Module Scheduling\\
in Dual-System Vision-Language-Action Models}
\author{Shilin Zhang\\
North China University of Technology\\
\texttt{zhangshilin@ncut.edu.cn}}
\date{October 2026}

\begin{document}
\newcommand{\proofref}{Appendix~\ref{app:proofs}}
\maketitle

\begin{abstract}
Dual-system vision-language-action (VLA) models pair a high-frequency
fast policy with a slow, more expensive module, and differ both in
\emph{how} the module intervenes (latent refreshing, value-guided
resampling, action-suffix reuse) and in \emph{when} it is invoked.
The invocation question has just begun to be mechanized by trigger
mechanisms, but these come with no shared objective, no measured
costs, and no evaluation standard that separates schedule quality
from slow-module quality. We provide the missing account. We
formalize invocation as a calibrated decision problem under a query
budget, show that fixed-ratio, fixed-budget, and history-gated
scheduling are degenerate cases of one quantity, the Value of
Intervention (VoI), and prove a renewal baseline that locates the
economics: under state-independent collapse the optimal schedule is
periodic, and the gain of any adaptive trigger is governed by the
cost ratio $\rho = c_{\text{slow}}/c_{\text{fast}}$ and the temporal
dispersion of collapse. Published systems span an order of magnitude
on $\rho$; measuring FiS-VLA on one RTX 4090
places it at the low end, one slow refresh costing $52$\,ms, about
one fast action ($55$\,ms) or $3.8$ denoising steps, which explains
why its hand-tuned $1{:}4$ cadence is reasonable and bounds what any
trigger can gain there. A direct test confirms the prediction: on a
controlled testbed the calibrated advantage over the fixed cadence
vanishes under homogeneous collapse and grows monotonically with
collapse heterogeneity. A calibrated router instantiates the account:
on a synthetic testbed with matched budgets it Pareto-dominates all
three baseline families ($0.825\pm0.002$ vs.\ $0.762\pm0.002$ success
at a $0.25$ query rate), and we characterize its boundary, not only
its wins: the advantage is insensitive to estimator variance but
vanishes once the estimator loses validity on most steps, the
requirement any trigger mechanism must meet to pay. Finally, we
contribute a four-measure protocol with an explicit leak-free rule;
on 929 deployment traces from RoboICL the rule is load-bearing,
cutting the apparent early-warning AUC of schedule statistics from
$0.93$ post hoc to $0.62$ ($p=0.010$), and failure attribution
reverses across task categories, so no single fixed ratio can exploit
the signal. A detection-limited frontier theorem converts such
early-warning AUC into a schedulability statement: rescoring
$33{,}443$ RoboICL decision points leak-free, a threshold router
reaches the deployed reuse gate's $33$--$48\%$ call-reduction regime
at an explicit missed-failure price, and the release's reported API
costs anchor the continuum's high-asymmetry end at $\rho \approx
10^{3}$, three orders of magnitude above FiS. We close the loop on
the released system itself:
fine-tuning FiS-VLA on RLBench, we measure the freshness value
function of the slow latent directly and find it nonpositive over
ages up to eight fast steps while amortized latency falls
$2.1\times$; the latent drifts with the $\sqrt{\delta}$ law of a
Wiener process, yet the action map is locally insensitive to it; and
a negative closed-loop result, realized under two fine-tuning
regimes, isolates the failure mechanism and yields a
competence-conditioned criterion for when success-level scheduling
comparisons are informative at all. A compilation loop completes the
account: on controlled task families the per-task period compiled
from the measured value function tracks the closed-loop oracle
within $2.4\%$, and the released checkpoint compiles to the boundary
its nonpositive value function dictates. Closed-loop arms on LIBERO
instantiate the account where the economics are contested
($\rho = 1.53$): the adaptive trigger allocates its queries exactly
as the threshold rule predicts, yet the language-rewrite channel
cannot convert them into success on either of two task suites,
while a revert actuator that undoes a failed rewrite is the only
intervention with zero net downside on both LIBERO-Spatial and
LIBERO-Goal, where the rewrite deficit deepens from thirteen to
forty-two points. Together these results turn slow-module
invocation from a design opinion into a measured decision.
\end{abstract}

\section{Introduction}
\label{sec:intro}

Vision-language-action (VLA) systems increasingly adopt a dual-system
architecture: a fast policy executes at control frequency, while a
slower, more expensive module intervenes occasionally to correct or
re-plan. This pattern appears in trained systems, where the slow layer
is a large VLM conditioned on the fast policy's latents
(Fast-in-Slow~\cite{fis2025}) or performs value-guided best-of-$N$
resampling (Hume~\cite{hume2025}), and in zero-training systems built
on frozen LLMs, where a gate decides when to reuse predicted action
suffixes (RoboICL~\cite{roboicl2026}).

These systems differ in \emph{how} the slow module acts. Our starting
observation is that they also differ in \emph{when} it acts, and that
this second choice has just begun to be mechanized: OneTwoVLA learns a
decision token that toggles between reasoning and acting; the
concurrent TUD trigger invokes the generalist on high predictive
uncertainty~\cite{yang2026tud}; System Switch
defers to a reasoning model on low
confidence~\cite{bailo2026systemswitch}. What these mechanisms lack
is an account of the problem they solve: no shared objective, no
measured costs, no evaluation standard that separates schedule
quality from slow-module quality, and no prediction for when
adaptivity beats a hand-tuned fixed ratio at all.

\begin{figure*}[t]
\centering
\includegraphics[width=\linewidth]{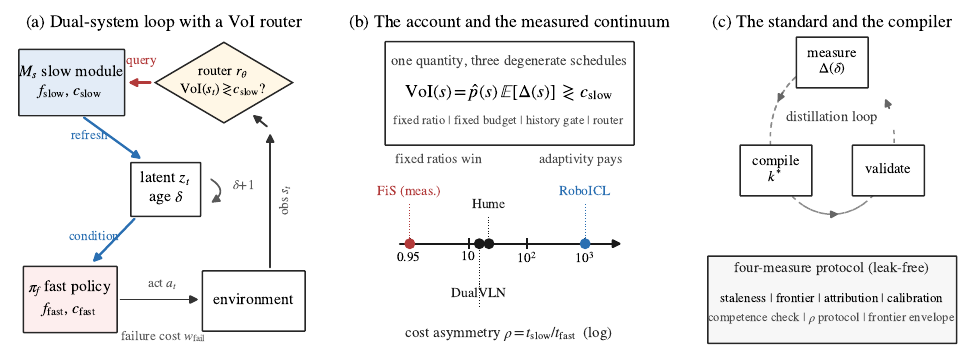}
\caption{Overview. (a)~The dual-system loop: a fast policy $\pi_f$
acts at control rate conditioned on a slow latent that ages between
refreshes; a router invokes the slow module $M_s$ only when the
Value of Intervention exceeds the query cost. (b)~The account: one
threshold rule subsumes fixed-ratio, fixed-budget, and history-gated
scheduling as degenerate cases, and the cost-asymmetry continuum
$\rho$ spans three orders of magnitude with both ends measured
(FiS on hardware, RoboICL from reported API costs). (c)~The
standard: a four-measure leak-free protocol and a distillation loop
that compiles the measured staleness value function into per-task
refresh periods.}
\label{fig:architecture}
\end{figure*}

\paragraph{Thesis.}
We give that account
(Figure~\ref{fig:architecture}). We model slow-module invocation as a
calibrated decision problem under a query budget; show that existing
fixed schedules are degenerate cases of one quantity, the \emph{Value
of Intervention} (VoI); and locate the economics of the problem on a
\emph{cost-asymmetry continuum} $\rho = c_{\text{slow}}/c_{\text{fast}}$
that spans three orders of magnitude across published systems and
that we measure at both ends ($\rho = 0.95$ on FiS-VLA hardware,
${\sim}10^{3}$ from the RoboICL deployment's reported API costs). This
yields four contributions.

\begin{enumerate}
\item \textbf{A decision-theoretic account with a cost baseline.} We
define VoI as the expected quality gap between the slow-corrected and
fast-only actions, weighted by the probability that the fast policy
collapses in state $s$ (Section~\ref{sec:method}). Fixed-ratio,
fixed-budget, and history-gated scheduling are degenerate cases of VoI
routing (Propositions~\ref{prop:fis}--\ref{prop:jev}). A renewal
baseline (Proposition~\ref{prop:continuum}) then locates the regimes:
under state-independent collapse the optimal schedule is periodic with
interval $\sqrt{2c_{\text{slow}}/(w_{\text{fail}}\pi)}$, so fixed
ratios are hard to beat at low $\rho$ and low collapse heterogeneity,
while adaptivity pays as either grows. This explains why hand-tuned
cadences are reasonable on embedded-latent systems and why triggered
invocation concentrates on separate-VLM systems.
\item \textbf{A calibrated router with its boundary characterized.} A
lightweight router estimates VoI online and queries the slow module
only when the estimate exceeds the query cost. We characterize where
this construction breaks, not only where it works: the advantage is
insensitive to estimator variance but vanishes when the estimator
loses validity on most steps, which is the requirement any trigger
mechanism must meet to pay (Sections~\ref{sec:exp-noise}--\ref{sec:exp-breakdown}).
\item \textbf{A leak-free measurement standard.} We contribute
staleness curves, routing frontiers, failure attribution, and
calibration reliability as a four-measure battery for slow-query
policies (Section~\ref{sec:protocol}), with an explicit leak-free
rule. On 929 deployment traces from the RoboICL release the rule is not decoration:
the same features reach AUC $0.93$ post hoc but $0.62$ ($p=0.010$)
leak-free, so without it the early-warning capability of schedule
statistics would be overstated by roughly $1.5\times$. The traces
also fix the precision of frontier claims themselves: skip decisions
cluster by episode at design effects of $8$--$18$, so a frontier
estimate over $16{,}000$ decision points carries the information of
about one thousand episodes
(Proposition~\ref{prop:dispersion}).
\item \textbf{Schedule compilation.} The measured quantities are not
only diagnostic. A renewal compiler turns a task's measured staleness
value function into a per-task refresh period; on controlled task
families the compiled period tracks the closed-loop oracle at a cost
gap of at most $2.4\%$, the compilation is a sample-average
approximation whose optimality gap admits a statistical
certificate~\cite{mak1999}, and the underlying trade-off inherits
the cost flatness of the Harris--Wilson order-quantity
problem~\cite{harris1913}, so moderate cost-estimate errors degrade
the achieved schedule only at second order; and the validation
budget itself has an optimal allocation, concentrated on the
near-ties of the cost bowl, which roughly doubles the probability
of recovering the oracle period at a fixed budget
(Remark~\ref{rem:ocba}). On the FiS backbone the
compiler sends every task to the boundary its measured (nonpositive)
value function dictates (Section~\ref{sec:exp-distill}). Adaptive
triggers, fixed periods, and compiled schedules become commensurable
as different amounts of measurement invested in the same cost
functional.
\end{enumerate}

\paragraph{Evidence.}
On a synthetic testbed with matched query budgets, calibrated routing
dominates the Pareto frontier of the fixed-ratio, always-query, and
reuse-gate baselines
(Section~\ref{sec:exp-synthetic}). On real systems, we apply the
protocol to all 929 deployment traces released by RoboICL
(Section~\ref{sec:exp-roboicl}): failed episodes query the slow module
$25\%$ less densely and run with greater staleness, and the effect is
category-dependent, so no single fixed ratio can exploit it. On
hardware we instantiate the cost side of the continuum with
FiS-VLA: one slow refresh costs $52$\,ms, about one fast action
($55$\,ms), or $3.8$ denoising steps, on a single RTX 4090
(Section~\ref{sec:exp-latency}), placing embedded-latent dual
systems at the $\rho \approx 1$ end, opposite the separate-VLM
systems at $\rho$ of $10$--$20$ that trigger-based invocation
targets, and the frozen-LLM API systems at $\rho \approx 10^{3}$
measured from the same release's reported costs
(Section~\ref{sec:exp-e32}): three orders of magnitude with both
ends anchored on hardware measurement or audited cost reports
(Table~\ref{tab:rho}). A closed-loop dual system placed at the
continuum's middle ($\rho = 1.53$: OpenVLA-7B as the fast policy and
Qwen2.5-VL-7B as the slow module on LIBERO-Spatial) then tests the
account where the economics are contested. Fourteen closed-loop arms
at matched seeds, initial states and budgets, eleven on LIBERO-Spatial
and a three-arm replication on LIBERO-Goal with an LLM slow endpoint,
sweep the two layers the
account separates: the adaptive trigger allocates its queries
exactly as the threshold rule predicts, yet a subgoal-rewrite
channel cannot convert them into success, and neither can the
screened, faster or slower variants of that channel, whose pooled
rate sits at $43\%$ against fast-only's $58\%$ with no dose axis in
sight. The failure is the channel's, not the schedule's: replacing
the language channel with action ranking, or adding an actuator that
undoes an unhelpful rewrite, changes the outcome's risk profile
rather than its mean, and the revert arm produces the table's only
direction-consistent positive contrast ($15$ vs.\ $4$ discordant
pairs, nominal $p = 0.019$, null under family correction), landing
on exactly the tasks the risk coordinate predicts. The cross-suite
replication keeps the ordering: the rewrite channel's deficit
deepens to $42$ points on the second suite while the revert arm
tracks fast-only exactly ($48/100$ against $48/100$). The operating
point is thus a boundary case in which the deployment, not the
module, sets the value of asking
(Section~\ref{sec:exp-libero}).

\paragraph{Scope.}
This is a methodology contribution: a decision-theoretic account, a
constructive router with characterized boundaries, and a measurement
standard. Third-party systems enter only as measurement subjects:
they anchor the continuum's two ends and stress-test the standard;
no contribution depends on their internals or on access beyond
public artifacts. The account's low-$\rho$ prediction is tested directly: on
the testbed, the calibrated advantage vanishes under homogeneous
collapse and grows monotonically with collapse heterogeneity, exactly
as the renewal baseline predicts
(Section~\ref{sec:exp-prediction}). Fine-tuning the open checkpoint
on four consumer GPUs and measuring its staleness frontier directly
confirms the prediction's transfer to the deployed FiS system
(Section~\ref{sec:exp-frontier}).

\section{Related Work: Three Families, One Design Space}
\label{sec:related}

We group dual-system VLA work by the mechanism that connects the fast
policy to the slow module, and summarize the comparison in
Table~\ref{tab:taxonomy}.

\begin{table*}[t]
\centering
\footnotesize
\setlength{\tabcolsep}{4pt}
\caption{Taxonomy of the source systems plus the concurrent triggers,
with the cost asymmetry $\rho$ of Eq.~\eqref{eq:rho} where it is
recoverable: measured on hardware for FiS, computed from published
control rates for Hume, from published API costs for RoboICL. No
trigger mechanism reports $\rho$, the quantity that decides whether
it can pay (Section~\ref{sec:continuum}). Entries are restricted to
claims verified against the papers' own sources.}
\label{tab:taxonomy}
\begin{tabular}{@{}lllllc@{}}
\toprule
System & Family & Training & Slow mechanism & Scheduling & $\rho$ \\
\midrule
FiS-VLA~\cite{fis2025} & trained & SFT + RL & latent refresh & fixed $1{:}4$ & $0.95$ (measured) \\
Hume~\cite{hume2025} & trained & Cal-QL head & value best-of-$N$ & fixed budget & ${\sim}22$ (from rates) \\
RoboICL~\cite{roboicl2026} & frozen LLM & none & suffix reuse gate & history heuristic & ${\sim}10^{3}$ (from costs) \\
TUD~\cite{yang2026tud} & trained & none & generalist re-call & uncertainty trigger & not reported \\
System Switch~\cite{bailo2026systemswitch} & trained & none & confidence defer & confidence trigger & not reported \\
\bottomrule
\end{tabular}
\end{table*}

\paragraph{Latent refreshing.}
Fast-in-Slow~\cite{fis2025} trains a single 7B backbone whose slow
path conditions the fast path through latents refreshed every $k$
steps; the ratio $1{:}4$ is selected by ablation over $1{:}1$ to
$1{:}8$. The schedule is fixed at deployment regardless of task
difficulty. We emphasize a nuance for later comparisons: the slow
layer here is a \emph{refresh cadence} on a latent condition, not an
independent low-frequency process; the reported $21.9$\,Hz control
rate is the whole system, and the effective slow refresh tick is
$21.9/4\approx5.5$\,Hz.

\paragraph{Value-guided resampling.}
Hume~\cite{hume2025} pairs a VLM System-2 ($4$\,Hz) with a lightweight
diffusion System-1 ($90$\,Hz) and samples $N$ action candidates,
selecting by a Cal-QL-trained value head. The schedule question
becomes the budget $N$, which is fixed. The value head is the closest
published antecedent of our calibration measure, but it is trained per
system and used for selection, not for deciding \emph{whether} to
consult the slow module at all.

\paragraph{Reuse gating on frozen LLMs.}
RoboICL~\cite{roboicl2026} shows that a frozen LLM can act as a robot
policy via typed in-context programs. Its Jev gate suppresses
redundant LLM calls by reusing predicted action suffixes, cutting
calls by $33$--$48\%$. Two facts from the source matter here. First,
the main benchmarks run \emph{without} the Jev gate; it is an
auxiliary study. Second, the gate is history-based and its task
success is uneven (it degrades some benchmarks), which we analyze as a
missing-calibration failure mode.

\paragraph{Uncertainty-triggered invocation (concurrent).}
While this paper was in preparation, the scheduling question has begun
to be mechanized. OneTwoVLA learns a decision token that toggles the
system between reasoning and acting modes, at the cost of manually
annotated phase boundaries~\cite{yang2026tud}. The concurrent TUD
trigger measures the cross-step dispersion of the specialist's action
re-predictions under the cached generalist context and invokes the
generalist when dispersion crosses a threshold, reporting higher
success at matched call budgets and a $75\%$ call reduction on a real
robot~\cite{yang2026tud}. System Switch defers decisions from a fast
actor to a reasoning VLM on low confidence and shows the gain tracks
the actor's calibration quality~\cite{bailo2026systemswitch}. These
mechanisms are complementary to our contribution rather than
competing with it: each is a particular estimator of the
collapse-probability factor of VoI, deployed at the high-asymmetry
end of the continuum (separate VLM generalists), and none reports the
measured query costs, the collapse heterogeneity, or the leak-free
evaluation that determine whether a trigger can pay
(Section~\ref{sec:continuum}).

\paragraph{Status-update theory.}
The scheduling question we pose has a classical upstream in the
age-of-information and remote-estimation literature: a monitor must
decide when to pay for a fresh sample of a drifting
signal~\cite{yates2019aoi,yates2021aoisurvey}. Sun, Polyanskiy, and
Uysal prove that the optimal sampling policy for a Wiener process under
a sampling-rate constraint is a threshold rule whose period scales with
the square root of the cost-to-drift ratio~\cite{sun2020wiener}, and
\cite{sun2017update} shows the same trade-off in the
generate-at-will model, where eager sampling can be strictly worse
than waiting. These results are derived for scalar signals and
communication channels; none has been applied to embodied systems,
where the signal is a learned latent, the cost is wall-clock latency of
a 7B model, and the consumer is a diffusion policy. Our contribution
relative to this literature is transport and measurement:
Proposition~\ref{prop:continuum} is the Bernoulli-collapse instance of
the same family, and Section~\ref{sec:exp-frontier} measures both
factors of the threshold rule, query cost and latent drift, on the
FiS-VLA system rather than assuming them.

\paragraph{Caching.}
Web-caching theory is a third upstream that prices the same object:
a TTL-based cache holds a stale copy and pays an age-dependent miss
penalty, the cache-side instance of the freshness value $V$; Cohen
and Kaplan quantify how miss rates depend on the age of cached
copies~\cite{cohen2001,cohen2005}. What the caching and
status-update literatures leave open, and what the continuum
addresses, is the regime where the refresh is not a disk read but a
multi-second VLM call, three orders of magnitude past the cost
asymmetries those literatures were built for.

\paragraph{Positioning.}
Adaptive computation in language models reduces cost within a single
model by early exiting~\cite{teerapittayanon2016branchynet}, adaptive
depth~\cite{elbayad2020depthadaptive}, or token
pruning~\cite{kim2022tokenpruning}, while action
chunking~\cite{zhao2023act} amortizes policy calls over predicted
horizons. These address budget questions inside one model; our
question is the budget for \emph{consulting a separate slow module}
inside a dual-system VLA. Calibrated confidence
estimation~\cite{guo2017calibration} underpins our collapse estimator,
and offline RL value heads~\cite{nakamoto2023calql} are the closest
published antecedent of the calibration signal we rely on, though they
are trained for selection rather than for scheduling. Prior and
concurrent work thus supplies trigger \emph{mechanisms}; none supplies
the decision-theoretic account that subsumes the fixed schedules these
mechanisms improve on, the measured cost economics that decides
whether any mechanism can pay, or a leak-free standard for comparing
them across system families. That account and standard are what we
contribute.

\section{Method: Routing as a Calibrated Decision Problem}
\label{sec:method}

\subsection{Setup}
Let $\pi_f$ be a fast policy executed at every control step $t$, and
$M_s$ a slow module that, given a query at time $t$, returns a
correction $c_t$ (a refreshed latent, a resampled action, or a
re-plan). A \emph{schedule} is a (possibly randomized) map
$r_t: \mathcal{H}_t \to \{0,1\}$ deciding whether to query $M_s$,
where $\mathcal{H}_t$ is the information available at step $t$.
Querying costs $c_{\text{slow}}>0$; task failure costs
$w_{\text{fail}}>0$. The scheduler's objective is
\begin{equation}
\min_{r}\;\; \mathbb{E}\Big[\, w_{\text{fail}}\,\mathbf{1}\{\text{fail}\}
\;+\; c_{\text{slow}} \textstyle\sum_t r_t \,\Big].
\label{eq:objective}
\end{equation}

\subsection{Value of Intervention}
\begin{definition}[Collapse and VoI]
\label{def:voi}
Let $Q_f(s)$ denote the task-value attained by continuing with the
fast policy alone from state $s$, and $Q^*(s)$ the value attained with
one slow correction followed by optimal continuation. The fast policy
\emph{collapses} in $s$ when the gap $\Delta(s) = Q^*(s)-Q_f(s)$
exceeds a margin $\epsilon \ge 0$, with indicator
$\xi_\epsilon(s) = \mathbf{1}\{\Delta(s) > \epsilon\}$. The
\emph{Value of Intervention} is
\begin{equation}
\begin{split}
\mathrm{VoI}(s) \;=\; \mathbb{E}\big[\,\Delta(s)\,\big|\,s\big]
\;=\; \mathbb{E}\big[\Delta(s) \mid \xi_\epsilon(s){=}1,\, s\big]\\
\times\; P(\xi_\epsilon(s){=}1 \mid s).
\end{split}
\label{eq:voi}
\end{equation}
\end{definition}

The optimal schedule under~\eqref{eq:objective} queries exactly when
$\mathrm{VoI}(s_t) > c_{\text{slow}}$ (\proofref).
VoI factorizes into a magnitude term and a probability term; the three
source systems implicitly fix one of the two.

\subsection{Three Baselines as Degenerate Cases}
\begin{proposition}[Fixed ratio \emph{(FiS)}]
\label{prop:fis}
A schedule that queries every $k$-th step is VoI-optimal iff
$\mathrm{VoI}(s_t)$ is constant across time and equals $c_{\text{slow}}$
at the query steps, i.e., iff the collapse probability is (wrongly)
assumed time-invariant.
\end{proposition}

\begin{proposition}[Fixed budget \emph{(Hume)}]
\label{prop:hume}
A schedule that queries everywhere with a fixed best-of-$N$ budget is
VoI-optimal iff $P(\xi_\epsilon(s_t){=}1\mid s_t)$ is bounded below by
$c_{\text{slow}}/\mathbb{E}[\Delta \mid \xi{=}1]$ at every step, i.e.,
iff collapse is assumed pervasive. It overspends wherever the fast
policy is reliable.
\end{proposition}

\begin{proposition}[History gate \emph{(Jev)}]
\label{prop:jev}
Any schedule $r_t = g(\text{schedule history})$ that is measurable
with respect to past scheduling decisions alone (not the current
state's uncertainty) is suboptimal on any task distribution whose
collapse process is state-dependent and independent of schedule
history, conditional on the state trajectory.
\end{proposition}

Proposition~\ref{prop:jev} is the formal statement of the failure we
measure empirically in Section~\ref{sec:exp-roboicl}: scheduling
history contains no early signal of collapse, so history-based gates
inherit exactly the variance that calibrated state estimates would
remove. Proofs are in \proofref.

\subsection{The Cost-Asymmetry Continuum}
\label{sec:continuum}

The propositions above characterize \emph{which} information a
schedule should use; none says how much a schedule stands to gain,
which depends on what a query costs. Let
$\rho = c_{\text{slow}}/c_{\text{fast}}$ denote the cost of one slow
query in fast-action units. Published dual systems span more than an
order of magnitude on this single number: a separate-VLM generalist
running at $2$--$4$\,Hz against a $30$--$90$\,Hz specialist puts
$\rho$ around $10$--$20$~\cite{hume2025,wei2025dualvln,yang2026tud},
whereas on FiS-VLA
hardware we measure $\rho = 0.95$
(Section~\ref{sec:exp-latency}): one latent refresh costs about one
fast action. The following baseline shows that this number, not the
choice of trigger mechanism, determines the economic regime.

\begin{proposition}[Cost-symmetric baseline]
\label{prop:continuum}
Suppose collapse follows a Bernoulli process with constant per-step
rate $\pi$, a collapse that begins persists until the next slow query,
and queries are periodic with interval $k$. The expected cost per step
is $C(k) = c_{\text{slow}}/k + w_{\text{fail}}\,\pi\,k/2$, minimized at
\begin{equation}
k^{*} \;=\; \sqrt{2\,c_{\text{slow}}/(w_{\text{fail}}\,\pi)}\,,
\label{eq:kstar}
\end{equation}
and under state-independent collapse the optimal schedule is
periodic. The advantage of any state-adaptive schedule over this
periodic baseline is zero when the collapse rate is time-invariant
and grows with its dispersion over time
(\proofref).
\end{proposition}

Three consequences locate the design regimes. First, at the
low-asymmetry end ($\rho \approx 1$) a fixed ratio is hard to beat:
queries are cheap relative to failure, so $k^{*}$ is large whenever
the fast policy is reliable, and the residual gain of adaptivity is
bounded by how unevenly collapse is distributed in time. This
explains, at measured cost, why the hand-tuned $1{:}4$ cadence of
FiS ($+24\%$ compute; Section~\ref{sec:exp-latency}) is a reasonable
operating point. Second, at the high-asymmetry end ($\rho \gg 1$),
every unnecessary query is expensive in fast-action units and
mis-scheduling is costly on both sides; this is where
uncertainty-triggered invocation pays~\cite{yang2026tud,bailo2026systemswitch}. Third, the two factors
of the threshold rule come apart exactly as the continuum predicts:
concurrent trigger mechanisms differ in how they estimate the
collapse factor, but whether a mechanism can pay depends on $\rho$
and on collapse heterogeneity, quantities that no trigger paper
currently reports. The four-measure protocol
(Section~\ref{sec:protocol}) makes both visible, and the controlled
sweeps of Sections~\ref{sec:exp-noise}--\ref{sec:exp-breakdown}
probe the heterogeneity axis directly.

The baseline model is deliberately minimal; its role is to locate
regimes, not to fit numbers. The persistence assumption is
optimistic for collapse that self-corrects, which shrinks the value
of querying at all; we return to this in
Section~\ref{sec:discussion}.

\begin{remark}[Drift instance of the same scaling]
\label{rem:drift}
The persistence model treats staleness cost as arising from discrete
collapse events. The complementary model treats the slow latent as a
continuously drifting status signal, in the sense of remote estimation
of a Wiener process~\cite{sun2020wiener}: if a latent aged $\delta$
steps incurs fidelity loss $\Delta(\delta) \approx a\,\delta^{2}/2$
(random walk with per-step intensity $\pi_d$), the renewal-reward
average cost $C(k) = c_{\text{slow}}/k + c_{\text{fast}} +
\frac{1}{k}\sum_{\delta=1}^{k}\Delta(\delta)$ is minimized at
$k^{*} = \sqrt{c_{\text{slow}}/a} = \sqrt{2c_{\text{slow}}/(w_{\text{fail}}\pi_d)}$,
the same square-root family as
Eq.~\eqref{eq:kstar} with the collapse rate $\pi$ replaced by the
drift intensity $\pi_d$. The scaling of the optimal period is thus
robust across the two canonical status-update models, event-collapse
and continuous drift; which model applies is an empirical question
about the trained system, and Section~\ref{sec:exp-frontier} answers
it by measuring the staleness-cost curve $\Delta(\delta)$ directly.
\end{remark}

\begin{proposition}[Sampling-class optimality]
\label{prop:sampling}
Retain the setup of Definition~\ref{def:voi} and let the scheduler
observe, between queries, a signal $z_t$ generated conditional on the
collapse state (proprioception, action statistics, scene change).
Then: (i) the schedule minimizing Eq.~\eqref{eq:objective} is a
threshold on the posterior collapse probability,
\begin{equation}
\label{eq:posterior}
\begin{split}
r_t &\;=\; \mathbf{1}\bigl\{\,P(\xi_\epsilon{=}1 \mid z_{1:t})
\;\ge\; \lambda^{*}\,\bigr\},\\
\lambda^{*} &\;=\;
\frac{c_{\text{slow}}}{\mathbb{E}[\Delta \mid \xi_\epsilon{=}1,\,
s_t]}\,,
\end{split}
\end{equation}
which is the Shiryaev rule of quickest detection with a query
cost~\cite{shiryaev1963}; (ii) a periodic schedule is optimal only in
the \emph{zero-evidence class}, where the posterior is invariant along
the trajectory because $z_t$ carries no information about future
collapse; (iii) the identifiable content of adaptivity is the
channel's information about collapse: two schedules at the same query
rate differ in expected failure cost only where their posteriors
differ from the trajectory mean, so an adaptive trigger's closed-loop
gain is a functional of that information and vanishes with it, and
the gain cannot be separated from the channel that carries it.
\end{proposition}

\begin{proof}[Proof sketch]
(i) substitutes the posterior odds into Eq.~\eqref{eq:voi}. The
magnitude factor is state-independent under the proposition's
assumption, so the threshold rule on $\mathrm{VoI}$ becomes a
threshold on the posterior; Shiryaev's optimality covers the case
where the query cost enters as a fixed charge per stop. (ii) If the
posterior is invariant, Eq.~\eqref{eq:posterior} fires with a
state-independent probability, and among schedules with a fixed
firing probability the periodic one minimizes the amortized query
cost at equal staleness exposure, which is the optimality condition
of Proposition~\ref{prop:continuum}. (iii) Two schedules at the same
query rate have the same expected query cost; their expected failure
costs can differ only where the posterior differs from its
trajectory mean, so the failure-cost gain of adaptivity is a
functional of the channel's information about collapse and vanishes
with it.
\end{proof}

Proposition~\ref{prop:sampling} sharpens the continuum's heterogeneity
condition into an information-theoretic one, and it turns the
question ``does a trigger help?'' into a measurable one: estimate how
much the between-query channel knows about collapse, and price it
against $\rho$. The status-update literature reaches the same
classification from the sampling side: for Wiener sources the optimal
sampler is threshold-structured on the estimation error rather than
periodic~\cite{sun2020wiener}, and periodic sampling is optimal only
for degenerate source dynamics. The LIBERO arms of
Section~\ref{sec:exp-libero} instantiate the classification directly:
the stall detector reads a between-query signal (end-effector
displacement), the periodic arm reads none, and at matched budget the
threshold arm concentrates its queries on exactly the episodes where
the fast policy has stalled.

\begin{remark}[Triggers as windowed sequential tests]
\label{rem:sprt}
Proposition~\ref{prop:sampling} says the optimal trigger reads
evidence; it does not say how much evidence to buy before acting. The
finite-sample answer is Wald's sequential probability ratio test: for
simple hypotheses on the collapse state, the likelihood-ratio rule
with boundaries $A < 1 < B$ attains type-I and type-II errors at or
below $\alpha$ and $\beta$ while minimizing the expected number of
observations, among all rules with those error
bounds~\cite{wald1947}. A deployed trigger is the windowed
approximation: a statistic of the last $w$ steps compared against a
fixed threshold. Its window and threshold select an operating point
on the same type-I/type-II plane that the SPRT optimizes over, and
the stall detector of Section~\ref{sec:exp-libero} ($w = 15$ control
steps, $2$\,cm displacement) is such an approximation, with the
honest gap that its statistic is a heuristic proxy for the likelihood
ratio rather than the likelihood ratio itself. The operational
reading is that the window is the evidence budget: shortening it
queries earlier at a higher false-query rate, lengthening it queries
later at a lower one, and the measured $\rho$ prices the trade-off
through Eq.~\eqref{eq:kstar}.
\end{remark}

\begin{remark}[Inventory-theoretic instance and cost insensitivity]
\label{rem:eoq}
The renewal objective is also the Harris--Wilson order-quantity
trade-off~\cite{harris1913,zipkin2000}: a fixed charge
$c_{\text{slow}}$ amortized over a cycle of $k$ fast steps against a
staleness cost that accumulates with the cycle. In the linear-decay
regime, $\overline{\Delta}(k) = \frac{1}{k}\sum_{\delta \le k}
a\,\delta = a(k{+}1)/2$, so the cycle-dependent part of $C$ is
exactly the economic-order-quantity cost
$c_{\text{slow}}/k + (a/2)\,k$, and the classical sensitivity
identity applies: evaluating at $\gamma\,k^{*}$ multiplies that part
by
\begin{equation}
\label{eq:eoq}
\frac{C_{\mathrm{eoq}}(\gamma\,k^{*})}{C_{\mathrm{eoq}}(k^{*})}
\;=\;\frac{1}{2}\Bigl(\gamma + \frac{1}{\gamma}\Bigr)
\;\le\;1 + \frac{(\gamma-1)^{2}}{2},
\end{equation}
while the additive constants ($c_{\text{fast}}$ and the $a/2$ offset
of the discrete sum) make the realized relative penalty smaller
still. Two consequences matter beyond the algebra. First, errors in
the measured inputs enter the compiled period at first order and the
achieved cost at second order: a two-fold mis-estimate of
$c_{\text{slow}}$ shifts $k^{*}$ by $\sqrt{2}$ but the cost by only
about $6\%$ of the cycle-dependent part, so compiled schedules
transfer across systems whose measured costs differ moderately.
Second, the flat bowl explains why measurement investment saturates:
once probe-driven uncertainty in $k^{*}$ is small relative to the
bowl, further probes buy little, which
Section~\ref{sec:exp-distill} observes directly and certifies
statistically.
\end{remark}

\begin{remark}[Optimal probe allocation]
\label{rem:ocba}
The validation stage of the measure--compile--validate loop is a
ranking-and-selection problem: $K$ candidate periods are compared by
probe rollouts whose per-episode costs are independent with means
$\mu_k$ and variances $\sigma_k^2$, and the compiler returns
$\hat{k} = \arg\min_k \hat{C}(k)$. The asymptotically optimal
computing budget allocation of~\cite{chen2000ocba} concentrates
replications where the decision is actually close: for the incumbent
best $b$ and suboptimality gaps $\Delta_k = \mu_k - \mu_b$,
\begin{equation}
\label{eq:ocba}
\frac{n_k}{n_j} \;=\;
\frac{(\sigma_k/\Delta_k)^{2}}{(\sigma_j/\Delta_j)^{2}},
\qquad
n_b \;=\; \sigma_b\Bigl(\textstyle\sum_{k \ne b}
n_k^{2}/\sigma_k^{2}\Bigr)^{1/2},
\end{equation}
which maximizes the probability of correct selection at any total
budget. The rule prices two intuitions that the flat-bowl remark
makes only qualitative. Budget follows near-ties: a candidate
$1\%$ worse than the optimum needs roughly $(\sigma/\Delta)^2$ times
the replications of one $10\%$ worse, so the closed-loop cost bowl's
width, not the size of the search grid, sets the measurement bill.
And the marginal value of probes collapses once the residual gaps
exceed the noise scale, which is the saturation that
Section~\ref{sec:exp-distill} measures and the EOQ identity
explains. Allocation by Eq.~\eqref{eq:ocba} is the prescriptive
counterpart: it tells a practitioner where the next probe step buys
selection accuracy and where it buys nothing.
\end{remark}

\begin{remark}[Competence-conditioned measurement]
\label{rem:attribution}
Closed-loop cost decomposes, to first order, into a term that
scheduling cannot influence and a term that only scheduling can: an
action-competence floor $B(\varepsilon, H)$, set by the per-step
action error $\varepsilon$ of the fast policy compounding over the
episode horizon $H$, plus the freshness value
$V(k) = \frac{1}{k}\sum_{\delta \le k}\Delta(\delta)$ that
Section~\ref{sec:exp-frontier} measures. A scheduler acts only on
$V$; the threshold rule compares $\mathrm{VoI}(s_t)$ with
$c_{\text{slow}}$ and has no handle on $B$. Two measurement
consequences follow. First, a closed-loop success-versus-$k$
comparison is informative only when the competence floor sits below
the freshness value scale: when $B$ saturates the metric, every
schedule fails and the comparison cannot detect $V$ even if
$V > 0$. Second, the two terms are confounded in observational
attributions, because failure modes determine episode structure and
episode structure determines staleness; the direction of the
confounding is system-dependent. The open-loop protocol of
Section~\ref{sec:protocol} is built around this decomposition: it
measures $V$ under teacher forcing, where $B$ is held fixed, and
the influence control (a shuffled-task latent) calibrates the scale
at which the latent matters at all. A closed-loop study that skips
the competence check attributes to scheduling whatever the policy
cannot do. The scheduler has no more handle on the interface's
transfer coefficient $\eta_{\mathcal{C}}$ of
Proposition~\ref{prop:channel} than on $B$: the realized freshness
value is $V\cdot\eta_{\mathcal{C}}$, not $V$.
\end{remark}

\begin{proposition}[Channel factorization of intervention value]
\label{prop:channel}
Retain the setup of Definition~\ref{def:voi} and let the interface
$\mathcal{C}$ between the slow module's output space and the fast
policy's action realize the correction only in part:
\begin{equation}
\label{eq:channel}
\Delta_{\mathcal{C}}(s) \;=\; \eta_{\mathcal{C}}(s)\,\Delta(s),
\qquad
\mathrm{VoI}_{\mathcal{C}}(s) \;=\; \eta_{\mathcal{C}}(s)\,
\mathrm{VoI}(s),
\end{equation}
where the \emph{signed transfer coefficient}
$\eta_{\mathcal{C}}(s)\in\mathbb{R}$ is the expected fraction of the
correction's value that survives the interface, signed by whether the
surviving part moves the fast policy's action toward or away from the
corrected behavior: $\eta = 1$ for an interface that executes the
correction verbatim, $\eta = 0$ for one that discards it, and
$\eta < 0$ for one that lands the correction but misdirects the
action. Then: (i) the schedule minimizing Eq.~\eqref{eq:objective}
queries iff $\eta_{\mathcal{C}}(s_t)\,\mathrm{VoI}(s_t) >
c_{\text{slow}}$, so the interface scales, and can reverse, the
threshold rule's decision boundary; (ii) two schedules with identical
trigger statistics, query rates and $\rho$ have opposite-signed total
value iff $\eta_{\mathcal{C}}$ changes sign between their interfaces,
and an aggregate closed-loop comparison of such schedules measures
the task-mixture of $\eta$, which can cancel to zero while $|\eta|$
is large; (iii) $\eta_{\mathcal{C}}(s)$ is a property of the
module--channel--task triple, so the continuum's prediction that
adaptive triggers gain over fixed ratios as $\rho$ grows is a claim
about the detection layer alone: it holds conditional on
$\eta_{\mathcal{C}} > 0$ and is unfalsifiable where the sign has not
been independently established.
\end{proposition}

\begin{proof}[Proof sketch]
Substituting Eq.~\eqref{eq:channel} into Eqs.~\eqref{eq:voi}
and~\eqref{eq:objective} multiplies the objective by
$\eta_{\mathcal{C}}$ at every state, which gives (i). Two schedules at
equal query rate have equal expected query cost, so their expected
cost difference is
$\mathbb{E}\big[(\eta_1-\eta_2)\,\Delta(s)\,
\mathbf{1}\{\xi_\epsilon(s){=}1\}\big]$, which changes sign with the
coefficient difference and cancels across tasks whenever the mixture
of $\eta$ over tasks crosses zero; this gives (ii). For (iii), every
detection result (Propositions~\ref{prop:continuum},
\ref{prop:frontier} and~\ref{prop:sampling}) is derived for $\Delta$,
not $\Delta_{\mathcal{C}}$; multiplying by a negative $\eta$ reverses
the inequality each result concludes while leaving the trigger
statistics that estimate it unchanged.
\end{proof}

Proposition~\ref{prop:channel} splits the routing problem into two
layers that have to be measured separately. The \emph{detection
layer}, when to query, is what Propositions~\ref{prop:continuum},
\ref{prop:frontier} and~\ref{prop:sampling} characterize and what the
frontier and dispersion measurements estimate; all of them condition
on the correction being executed, so none of them sees the channel.
The \emph{channel layer}, whether a query's value survives the
interface, plays the role of a noisy channel in the
communication-theoretic sense~\cite{shannon1948}: it has a transfer
coefficient, and the coefficient, not the source, bounds what arrives
at the action. The decomposition explains why a closed-loop aggregate
can be null while both layers are strongly non-null:
Section~\ref{sec:exp-libero} measures a detection layer that
allocates queries exactly as the threshold rule predicts and a
channel layer that is harmful on the tasks the fast policy has
already solved. It also reframes what an actuator study measures: the
taxonomy of interfaces (subgoal rewrite, verification gate,
action-layer critic) is a measurement program for $\eta_{\mathcal{C}}$,
and two deployments of the same slow module can differ in sign.

\begin{remark}[Channel-layer adaptivity]
\label{rem:channel-adaptive}
A channel can carry adaptivity of its own, independently of the
trigger. If the interface ranks candidates sampled from the fast
policy's own action distribution, the transfer coefficient scales
with that distribution's uncertainty: where the policy is confident
the candidates coincide and the channel is a no-op
($\eta_{\mathcal{C}} \to 0$), and where the policy is uncertain the
candidates separate and the channel selects among them. The
correction budget then concentrates on uncertain states with no
detection layer at all, which is the channel-layer analogue of
Proposition~\ref{prop:sampling}: adaptivity can be bought at the
interface instead of at the trigger, the two are complementary, and a
deployment may spend its complexity in either place. The LIBERO arms
of Section~\ref{sec:exp-libero} instantiate the trigger-layer
version; the critic arm instantiates the channel-layer one, where the
slow module never touches the language channel and acts purely as a
selector among the fast policy's own samples.
\end{remark}

\begin{remark}[Channel risk]
\label{rem:channel-risk}
The transfer coefficient prices the mean of the realized correction;
two interfaces with equal $\eta_{\mathcal{C}}$ can still differ in
what the correction does when it misfires. Write $\Delta_t$ for the
paired per-task success difference an interface realizes against the
fast policy alone. The negative part
$D_{\mathcal{C}} = \sum_t \min(\Delta_t, 0)$ is the channel's
\emph{downside profile}, a coordinate the mean does not see: an
interface can be mean-null while catastrophic on the tasks it corrupts,
and mean-null while safe. The profile is governed by how far the
intervention can move the fast policy from the behaviour distribution
it was validated on. An interface that replaces the instruction moves
the conditioning off that distribution, and its harm is bounded only
by the task structure, which is the \emph{corruption regime}; an
interface that selects among candidates sampled from the fast
policy's own action distribution is confined to that distribution's
support, and its worst case is the spread of the candidates
themselves, which is the \emph{selection regime}. Support containment
is thus a design axis for actuators, orthogonal to
$\eta_{\mathcal{C}}$: intervening inside the validated behaviour
trades upside for a truncated downside, and a risk-sensitive
deployment ranks channels on both coordinates rather than on the mean
alone.
\end{remark}

\begin{proposition}[Detection-limited scheduling frontier]
\label{prop:frontier}
Let each decision point carry a leak-free risk score $S$ for episode
failure with $\mathrm{AUC}(S) = a$, and let a router skip the slow
query whenever $S \le \tau$. Write $r = P(\text{skip})$ for the call
reduction and $m = P(\text{skip} \mid y{=}1)$ for the missed-failure
rate, where $y$ labels a failing episode. Then:
(i) every label-blind policy, including every fixed period $k$,
satisfies $m = r$, so periodic scheduling operates on the diagonal;
(ii) if $S$ is binormal with equal variances, the router attains
\begin{equation}
\label{eq:frontier}
r \;=\; \Phi\!\bigl(\Phi^{-1}(m) + \sqrt{2}\,\Phi^{-1}(a)\bigr),
\end{equation}
so its advantage over the diagonal is governed by a single statistic,
the standardized separation $d' = \sqrt{2}\,\Phi^{-1}(a)$; at
$a = 0.5$ the frontier collapses onto the diagonal and conditional
routing degenerates to periodic scheduling. Since practical scores are
not likelihood ratios, Eq.~\eqref{eq:frontier} is an envelope, and the
empirical frontier of a real router lies below it.
\end{proposition}

\begin{proof}[Proof sketch]
(i) holds because a label-blind skip rule is independent of $y$, so
$P(\text{skip} \mid y{=}1) = P(\text{skip})$. For (ii), in the
equal-variance binormal model $P(S \le \tau \mid y{=}0) = \Phi(z)$ with
$z = (\tau - \mu_0)/\sigma$, $P(S \le \tau \mid y{=}1) = \Phi(z - d')$,
and $a = \Phi(d'/\sqrt{2})$; eliminating $z$ yields
Eq.~\eqref{eq:frontier}.
\end{proof}

Proposition~\ref{prop:frontier} converts a calibration measurement
into a schedulability statement. Its operative content is that the
AUC is a sufficient statistic for the envelope: no property of the
score beyond its ranking accuracy enters Eq.~\eqref{eq:frontier}, so
one number pins down the entire achievable trade-off between call
reduction and missed failures and quantifies exactly how much a
conditional router can beat any fixed period. A trigger paper that
reports an AUC at a measured $\rho$ has thereby reported its
schedulable region, whatever mechanism produced the score.
Section~\ref{sec:exp-e32} measures both the AUC and the frontier on
deployment traces from the RoboICL release and finds the empirical curve close to, but
below, the envelope, as the proposition anticipates.

\begin{proposition}[Clustered certification of the frontier]
\label{prop:dispersion}
Let the skip indicator of decision $t$ in episode $e$ follow a
one-way random-effects model $X_{e,t} = \mu + u_e + \varepsilon_{e,t}$
with $\mathrm{Var}(u_e) = \sigma_b^2$,
$\mathrm{Var}(\varepsilon) = \sigma_w^2$, and intraclass correlation
$\kappa = \sigma_b^{2}/(\sigma_b^{2}+\sigma_w^{2})$, and let the
router's call reduction $r$ be estimated by the pooled fraction
$\hat{r}$ over $n$ decision points grouped in episodes of mean size
$\bar{T}$. Then
$\sqrt{n}\,(\hat{r} - \mu) \xrightarrow{d}
\mathcal{N}\bigl(0,\, \mu(1-\mu)\,\mathrm{DEFF}\bigr)$ with design
effect
\begin{equation}
\label{eq:deff}
\mathrm{DEFF} \;=\; 1 + (\bar{T}-1)\,\kappa,
\end{equation}
so treating decisions as independent Bernoulli trials understates
the uncertainty of $\hat{r}$ by the design effect of
\cite{kish1965}. Certifying a frontier bound $r \le \bar{r}$ with
margin $\delta = \bar{r} - \mu$ at confidence $1-\alpha$ requires
$n \;\ge\; z_{1-\alpha}^{2}\,\mu(1-\mu)\,\mathrm{DEFF}/\delta^{2}$
decision points, and the same correction applies to the
missed-failure rate $m$ restricted to failing episodes.
\end{proposition}

\begin{proof}[Proof sketch]
The indicators are binary, so their total variance is exactly
$\sigma_b^{2} + \sigma_w^{2} = \mu(1-\mu)$. Within episode $e$ the
indicators share the effect $u_e$, hence
$\mathrm{Var}(\bar{X}_e) = \sigma_b^{2} + \sigma_w^{2}/T_e$.
Averaging $E$ episode means of common size $\bar{T}$,
$\mathrm{Var}(\hat{r}) = (\sigma_b^{2} +
\sigma_w^{2}/\bar{T})/E$; substituting $n = E\bar{T}$ and
$\kappa = \sigma_b^{2}/\mu(1-\mu)$ gives exactly
$\mu(1-\mu)\,[1 + (\bar{T}-1)\kappa]/n$, and the normal limit is the
cluster-sampling CLT. The certification bound inverts the resulting
interval; the unequal-cluster case replaces $\bar{T}$ by the $T_0$
of the one-way ANOVA estimator.
\end{proof}

Proposition~\ref{prop:dispersion} matters operationally because
labels in deployment traces are episode-constant and the router's
features overlap within an episode, so $\kappa$ is close to its
upper bound and the decision count is a misleading sample size:
Section~\ref{sec:exp-e32} measures $\kappa \approx 0.91$--$0.99$ on
the RoboICL release, design effects of $8$--$18$, and effective
sample sizes near one thousand decisions rather than the nominal
sixteen thousand. The closed-form certificate then matches an
episode-level bootstrap to within a few percent, which makes
frontier claims auditable at the episode count that actually
carries information.

\begin{proposition}[Model-free fallback]
\label{prop:competitive}
Let the expected staleness loss of the fast path accrue at rate $h$
per step while the latent goes unrefreshed, over an interval of
unknown length chosen by the environment, and let a slow query cost
$c_{\text{slow}}$ and reset the accrual. Then the break-even rule,
query once the accrued loss reaches $c_{\text{slow}}$ (after
$c_{\text{slow}}/h$ stale steps), is $2$-competitive against the
clairvoyant optimum on every interval; the randomized rule of
\cite{karlin1994} attains $e/(e{-}1)$; and predictions of the
interval length lift performance along the consistency--robustness
family of \cite{purohit2018,weizhang2020}, degrading to the
randomized guarantee when predictions are uninformative.
\end{proposition}

\begin{proof}[Proof sketch]
The problem is continuous-time ski rental with rental rate $h$ and
purchase price $c_{\text{slow}}$: on an interval of length $T$ the
clairvoyant pays $\min\{h\,T,\,c_{\text{slow}}\}$, and the cited
guarantees for the rent-or-buy problem transfer
verbatim~\cite{karp1992,karlin1994}. Concave accrual
($\Delta(\delta)$ sublinear in $\delta$) reduces the rental side and
only improves the ratio.
\end{proof}

Proposition~\ref{prop:competitive} is the decision-side complement
of Proposition~\ref{prop:frontier}. The frontier says what a
detector of a given AUC can buy; the competitive bound says what
survives when there is nothing to detect with. Even with no
calibrated router, no measured value function, and no measured
$\rho$, a system that queries when its accrued staleness equals the
query cost stays within a constant factor of optimal, and learned
predictions plug in exactly where they help. The measured drift law
$\Delta(\delta) \approx 0.12\,\delta$ is linear, so the accrual
assumption holds with the measured slope
$h = 0.12\,w_{\text{fail}}$, and the guarantee is quantitative
rather than asymptotic.

\subsection{The Calibrated Router}
\label{sec:router}
The router $r_\theta$ implements the threshold rule
\begin{equation}
\label{eq:router}
\text{query } M_s \iff r_\theta(s_t) \;=\;
\hat{p}(s_t)\,\hat{\mu}(s_t) \;>\; c_{\text{slow}},
\end{equation}
where $\hat{p}$ estimates the collapse probability and $\hat{\mu}$
the expected quality gap between slow-corrected and fast-only
actions. The rule is mechanism-agnostic. This paper runs three
instantiations of increasing fidelity and specifies each exactly.

\emph{(i) Calibrated-uncertainty threshold} (synthetic testbed,
Sections~\ref{sec:exp-synthetic}--\ref{sec:exp-breakdown}). The score
is a calibrated uncertainty estimate $\hat u_t$ of the per-step
collapse propensity; the router queries when $\hat u_t > \tau$, with
$\tau$ swept to trace the frontier and selected to match budgets in
the comparisons. Estimator quality is controlled exogenously, by
additive noise of variance $\sigma_u^{2}$
(Section~\ref{sec:exp-noise}) and by rank-destroying corruption on a
fraction $p$ of steps (Section~\ref{sec:exp-breakdown}).

\emph{(ii) Logistic router on schedule features} (deployment traces,
Sections~\ref{sec:exp-roboicl}--\ref{sec:exp-e32}). The score is a
logistic regression over leak-free prefix features of the decision
history (dispatch mean, dispersion, short-horizon fraction, density,
request--turn rejection rate, last dispatch length); episode-grouped
cross-validation produces the scores, so no episode contributes to
its own training fold.

\emph{(iii) Proposed online router} (design). The full instantiation
is a two-head multilayer perceptron on fast-policy features
(penultimate representation, action entropy over the chunk,
fast/slow disagreement on a probe sample, task-phase embedding).
Head $\hat{p}$ trains with binary cross-entropy on collapse labels
plus a calibration term, with expected calibration error reported;
head $\hat{\mu}$ trains with squared error on measured quality gaps.
Training data are episodes with spurious and injected perturbations,
labeled by the measured freshness value function of
Section~\ref{sec:exp-frontier}. At a 256-dimensional feature vector
the two-head network is ${\sim}42$k parameters, one forward pass per
control step, well under a millisecond; training touches only
$r_\theta$, with $\pi_f$ and $M_s$ frozen published checkpoints.
Instantiation (iii) is validated in this paper only through its two
operating proxies, (i) and (ii); closing that gap on the deployed
FiS checkpoint is the remaining integration of
Section~\ref{sec:roadmap}.

\paragraph{Model-agnostic contract.}
Interactions with $\pi_f$, $M_s$, and $r_\theta$ go through a narrow
interface (reset, observe, propose, correct), so any open fast/slow
pair can be wrapped and closed models can be swapped behind the same
contract with response snapshots frozen for measurement runs.

\section{The Four-Measure Protocol}
\label{sec:protocol}

Slow-query policies are currently evaluated by task success alone,
which conflates schedule quality with slow-module quality. We propose
four measures, all computable from logged rollouts plus a schedule's
query decisions, and all reported for every experiment in this paper.

\begin{enumerate}
\item \textbf{Staleness curve.} Task success as a function of elapsed
steps since the last slow query. Surfaces the delay at which a
schedule's fast-only stretch begins to fail.
\item \textbf{Routing frontier.} Success rate against query rate,
traced by sweeping the router threshold. A schedule is preferable if
its frontier is below-and-left (less cost, less failure). Reports of
a single operating point are not comparable across systems; frontiers
are.
\item \textbf{Failure attribution.} Distribution of slow queries over
task phases and of failure onsets over staleness. Separates ``the
slow module is weak'' from ``the schedule calls it too late''.
\item \textbf{Calibration reliability.} Reliability diagram and AUC of
the collapse predictor, with an explicit leak-free evaluation rule:
predictors may use only information available at decision time.
Whole-episode statistics are reported separately as post-hoc
descriptors, never as predictive claims.
\end{enumerate}

The leak-free rule is not bureaucratic caution. In
Section~\ref{sec:exp-roboicl} the same feature family yields AUC
$0.93$ when computed over whole episodes and $0.62$ when computed
from the first few decisions; a protocol without the rule would have
credited schedule features with more than twice their actual
forecasting power. As trigger mechanisms for the same scheduling
question multiply~\cite{yang2026tud,bailo2026systemswitch}, the
battery is what makes them comparable: frontiers at measured query
costs, not call counts at unmeasured ones.

For use as a review and evaluation instrument, the battery, the
competence check of Remark~\ref{rem:attribution}, the $\rho$ cost
protocol of Eq.~\eqref{eq:rho}, and the frontier envelope of
Proposition~\ref{prop:frontier} compress into five admission
questions that an evaluation of a slow-query policy should answer
before any scheduling claim is made.

\subsection{A Five-Item Checklist}
\label{sec:checklist}

\begin{enumerate}
\item \textbf{Competence floor.} Is the fast policy's standalone
action error below the scale at which scheduling could matter?
When the competence floor saturates the metric, every schedule fails
identically and the comparison cannot detect the freshness value
even if that value is positive
(Section~\ref{sec:exp-frontier}, closed-loop status).
\item \textbf{Leak-free rule.} Do the predictor's features use only
information available at decision time? Whole-episode statistics
inflated early-warning AUC by roughly $1.5\times$ on deployment traces
($0.93$ post hoc vs.\ $0.62$ leak-free;
Section~\ref{sec:exp-roboicl}).
\item \textbf{Censoring control.} Is the staleness--outcome
correlation corrected for the schedule's own effect on episode
length? The residual confound is system-dependent in direction;
matched-prefix scoring is the minimal control
(Section~\ref{sec:exp-e32}).
\item \textbf{Cost protocol.} Is $\rho$ reported under
Eq.~\eqref{eq:rho}, end-to-end wall-clock on the deployment path?
Call counts at unmeasured costs are not comparable across systems
(Table~\ref{tab:rho}).
\item \textbf{Frontier, not operating point.} Is the trade-off
reported as a curve with the periodic diagonal and the envelope of
Proposition~\ref{prop:frontier} drawn? A single operating point
cannot show whether a mechanism's advantage exceeds what its AUC
entitles it to (Figure~\ref{fig:routing-e32}).
\end{enumerate}

\begin{remark}[Certifying closed-loop comparisons]
\label{rem:exact}
The four measures are estimated from finite rollouts, and a
scheduling claim is a comparison between estimates; exact inference
keeps the comparison honest. With $n$ episodes at true success $p$,
the Clopper--Pearson interval is the unique interval with coverage
$1-\alpha$ for every $p$, so reported half-widths are worst-case
rather than Gaussian approximations. Two arms are separated at level
$\alpha$ only if their exact intervals are disjoint, which at
$n = 30$ requires a gap of roughly $0.25$; the minimum detectable
gap at power $1-\beta$ is
\begin{equation}
\label{eq:mdd}
\delta_{\min} \;=\;
\bigl(z_{\alpha/2} + z_{\beta}\bigr)\,
\sqrt{2\,\bar{p}(1-\bar{p})/n}\,,
\end{equation}
so a study that expects a $0.1$ effect needs on the order of $200$
episodes per arm, not $30$. When arms share seeds and initial states,
the paired design removes the between-episode variance component and
the discordant-pair count drives the McNemar standard error, which is
the variance reduction the LIBERO arm comparison of
Section~\ref{sec:exp-libero} exploits; Eq.~\eqref{eq:mdd} is then
conservative, because pairing replaces the total variance by the
discordant-pair variance. Finally, a family of $K$ arm
contrasts reported from one study needs multiplicity control (we use
Holm): under the global null, the probability that at least one of
ten contrasts reaches $p < 0.05$ is $1 - 0.95^{10} \approx 0.40$.
\end{remark}

\section{Experiments}
\label{sec:experiments}

We validate the three components of the account in complementary
ways: (i) a controlled synthetic testbed isolates scheduling from
slow-module quality and characterizes where calibrated routing helps
and where it breaks; (ii) at the continuum's high-asymmetry end, the
protocol and the detection-limited frontier of
Proposition~\ref{prop:frontier} are stress-tested on all 929
deployment traces released by RoboICL, and the same release's
reported API costs anchor the far end of the continuum
(Section~\ref{sec:exp-e32}); (iii) on hardware, a latency
measurement of FiS-VLA instantiates
the low-asymmetry end of the continuum
(Section~\ref{sec:continuum})
(Section~\ref{sec:exp-latency}); (iv) a direct test of the
continuum's low-$\rho$ prediction on the testbed
(Section~\ref{sec:exp-prediction}); (v) a backbone-level
integration that fine-tunes FiS-VLA on consumer GPUs, measures the
freshness value function, the latent drift law, and the influence
control that the scheduling theory consumes, and discloses a
negative closed-loop result with its mechanism
(Section~\ref{sec:exp-frontier}); and (vi) a compilation loop that
turns the measured value function into per-task refresh periods,
validated closed-loop on controlled families at a cost gap of at
most $2.4\%$ from the oracle and shown to recover the deployed
system's boundary decision from its measured curve
(Section~\ref{sec:exp-distill}). Remaining integrations are discussed
in Section~\ref{sec:roadmap}.

\subsection{Synthetic Testbed: Calibrated Routing Dominates Fixed Schedules}
\label{sec:exp-synthetic}

The testbed simulates a fast policy whose per-step collapse probability
is state-dependent and bursty (failure begets failure), a slow oracle
that collapse-free resets the state at query cost, and the four
schedules: always-query (Hume-like upper anchor), fixed ratio $1{:}k$
(FiS-like, $k\in\{2,4,8\}$), the Jev-style reuse gate, and the
calibrated router at thresholds sweeping the frontier. Success is
averaged over 5 seeds; $95\%$ CIs are within $\pm0.003$ throughout.

\begin{table}[t]
\centering
\footnotesize
\caption{Synthetic testbed (5 seeds). At matched budget $0.25$, the
calibrated router outperforms the FiS-like fixed ratio by $6.4$ points;
at half that budget ($0.127$) it still matches the full-budget fixed
$1{:}4$ schedule. The calibrated operating point at $0.202$ dominates
the Jev-style gate on both axes.}
\label{tab:synthetic}
\begin{tabular}{@{}llcc@{}}
\toprule
Schedule & Query rate & Success rate & $95\%$ CI \\
\midrule
Always query (Hume-like) & 1.000 & 0.921 & $\pm.001$ \\
Fixed $1{:}2$ & 0.500 & 0.815 & $\pm.002$ \\
Fixed $1{:}4$ (FiS-like) & 0.250 & 0.762 & $\pm.002$ \\
Fixed $1{:}8$ & 0.127 & 0.736 & $\pm.003$ \\
Jev-style gate & 0.222 & 0.778 & $\pm.002$ \\
\midrule
Calibrated, $c{=}0.3$ & 0.282 & \textbf{0.832} & $\pm.002$ \\
Calibrated, $c{=}0.4$ & 0.259 & 0.830 & $\pm.002$ \\
Calibrated, $c{=}0.5$ & 0.247 & 0.825 & $\pm.002$ \\
Calibrated, $c{=}0.6$ & 0.202 & 0.806 & $\pm.002$ \\
Calibrated, $c{=}0.7$ & 0.127 & 0.772 & $\pm.002$ \\
\bottomrule
\end{tabular}
\end{table}

\begin{figure}[t]
\centering
\includegraphics[width=\linewidth]{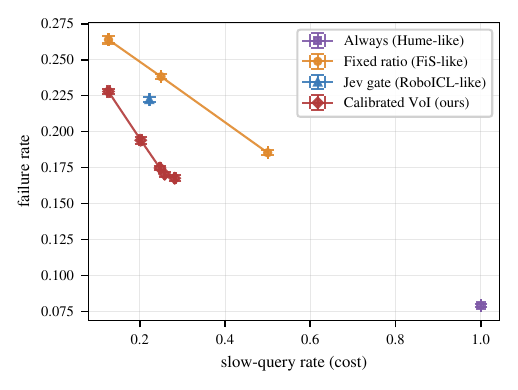}
\caption{Routing frontier (success vs.\ query rate, mean $\pm$ CI over
5 seeds). The calibrated frontier (red) sits below-and-left of the
fixed-ratio frontier (orange) across the budget range.}
\label{fig:frontier}
\end{figure}

Three observations. First, at the FiS operating budget ($q{=}0.25$),
calibrated routing gains $+6.4$ points; the gap widens as budgets
tighten, which is where scheduling matters most. Second, the Jev-style
gate is strictly dominated by a calibrated point with a \emph{lower}
query rate ($0.202$ vs.\ $0.222$) and higher success, consistent with
Proposition~\ref{prop:jev}. Third, staleness curves (in the released
artifacts) show the calibrated router concentrates queries in bursty
segments, explaining why its curve degrades more slowly than fixed
ratios as staleness grows.

\subsection{How Much Signal Quality Does Calibration Need?}
\label{sec:exp-noise}

\begin{figure*}[t]
\centering
\includegraphics[width=\textwidth]{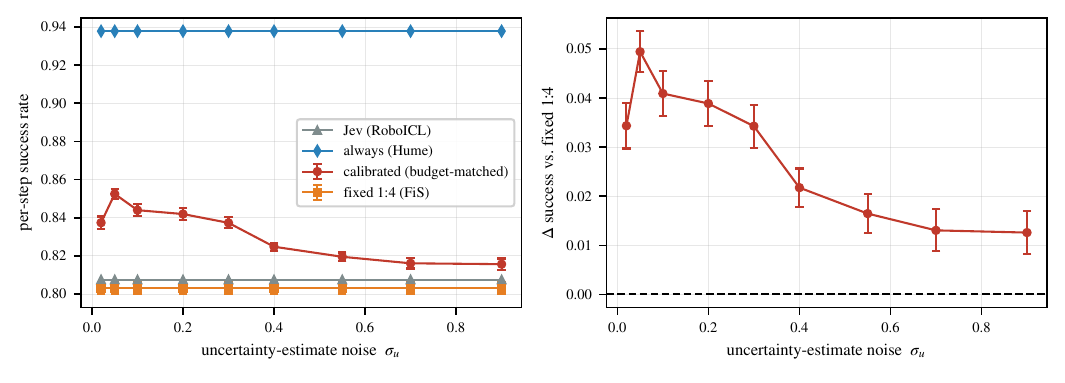}
\caption{Left: budget-matched success versus uncertainty-signal quality.
Right: advantage of calibrated routing over the fixed 1:4 schedule; it
decays monotonically but remains positive throughout.}
\label{fig:breakdown}
\end{figure*}

A calibrated router is only useful if its uncertainty input is
informative, so Section~\ref{sec:exp-synthetic} invites an obvious
objection: the advantage may be an artefact of a nearly noiseless
signal. We therefore sweep the quality of the uncertainty estimate,
$\hat u = \mathrm{clip}(d + \varepsilon)$,
$\varepsilon \sim \mathcal N(0, \sigma_u^2)$ over true difficulty $d$,
and re-compare at \emph{matched} slow-query budget. For each
$\sigma_u$ we fix the budget at the value realised by the 1:4 schedule
($q = 0.250$) and select the calibrated threshold whose realised query
rate is closest to it, so neither side wins by spending more. The
budget-matched comparison is reported in
Figure~\ref{fig:breakdown} rather than a table, since the figure
already carries every grid point: success falls only from $0.852$ to
$0.816$ as $\sigma_u$ grows from $0.05$ to $0.90$ at a realised query
rate held near $q = 0.25$.

The advantage decays gracefully and monotonically from $+0.049$ at
$\sigma_u = 0.05$ to $+0.013$ at $\sigma_u = 0.90$, and never crosses
zero (Figure~\ref{fig:breakdown}). Two caveats keep this honest. First, the
non-monotonicity at small $\sigma_u$ is a budget-matching artefact: at
$\sigma_u = 0.02$ the closest threshold realises $q = 0.172$, so the
calibrated router wins while spending \emph{less} than the fixed
schedule. Second, and more important, additive noise preserves the rank
correlation between $\hat u$ and $d$ for any $\sigma_u$, so this sweep
does not probe \emph{systematic} miscalibration: a biased or
rank-inverting estimator is a different failure mode, and testing it
requires the real-observation experiments of
Section~\ref{sec:roadmap}. What the sweep does establish is that the
routing rule is not fragile to estimator variance, which is the
objection most naturally raised against a threshold on a learned
uncertainty.

\subsection{Where Calibration Actually Breaks}
\label{sec:exp-breakdown}

The sweep above is weaker than it looks. Additive noise preserves the
rank correlation between $\hat u$ and $d$, so a threshold can always be
re-tuned to compensate, which is exactly why the advantage never
vanished for any $\sigma_u$. The failure mode that matters in
deployment is an estimator that is confidently \emph{wrong} on a random
subset of steps, which is what probe disagreement or action entropy
look like when they misfire. We model this directly: with probability
$p$ the estimate is replaced by an independent draw $\hat u \sim
U(0,1)$, destroying rank information on those steps; otherwise
$\hat u = \mathrm{clip}(d + \varepsilon)$ with $\sigma_u = 0.10$.
$p = 0$ reproduces Section~\ref{sec:exp-synthetic} and $p = 1$ reduces
the calibrated router to querying at random, every operating point
re-tuned to the matched budget as above. The advantage is flat up to
$p \approx 0.3$ ($+0.041$ at $p = 0$,
$+0.034$ at $p = 0.3$, at matched budgets of $q \approx 0.22$--$0.26$),
decays through the middle of the range, and is statistically
indistinguishable from zero from $p \approx 0.8$ onward ($+0.006$ at
$p = 0.7$ and $p = 0.8$, $-0.003$ at $p = 0.9$), reaching the
random-query limit at $p = 1$ ($-0.001$). So calibrated
routing tolerates an uncertainty estimator that is pure noise on up to
roughly half of all steps before the benefit is seriously eroded. This
is the honest version of the robustness claim: the rule is insensitive
to estimator \emph{variance} (Section~\ref{sec:exp-noise}) but does
require estimator \emph{validity} on a majority of steps.

A structural consequence is worth stating because it sharpens the
comparison to the baselines. Fixed-ratio, always-on and reactive gates
never consult $\hat u$, so their performance is invariant to $p$:
equally unaffected by signal quality and equally unable to exploit
it. Their robustness is a symptom of ignoring the signal; the
calibrated router's sensitivity to $p$ is the price of using it, and
the crossover near $p \approx 0.8$ is where that price exceeds the
benefit.

\subsection{Deployment Traces at the High-Asymmetry End}
\label{sec:exp-roboicl}

The RoboICL release includes decision-level traces (25\,Hz control; each decision
records its step and the number of fast steps dispatched since the
last slow query) for 929 evaluation episodes across four task
categories. We apply the protocol to all of them. To our knowledge
this is the first measurement of slow-query scheduling statistics on
a released trace set of a real dual-system LLM policy at this scale.
\begin{table*}[t]
\centering
\footnotesize
\caption{RoboICL traces: slow-query scheduling by outcome ($95\%$ CI).
Failed episodes query the slow module less densely and run with
greater staleness; the per-episode staleness gap ($+2.5$ steps) is
significant under label permutation ($p = 0.0005$).}
\label{tab:roboicl}
\begin{tabular}{@{}lcc@{}}
\toprule
Per-episode statistic & Success ($n{=}486$) & Failure ($n{=}443$) \\
\midrule
Slow decisions (count) & $39.4 \pm 1.5$ & $69.5 \pm 1.1$ \\
Decision density (queries/step) & $0.141 \pm .006$ & $0.105 \pm .005$ \\
Mean dispatched steps (staleness) & $9.8 \pm 0.5$ & $12.3 \pm 0.5$ \\
Late-decision fraction (last 20\%) & $0.183 \pm .002$ & $0.193 \pm .001$ \\
\bottomrule
\end{tabular}
\end{table*}

\paragraph{Failure attribution is category-dependent.}
Disaggregating by task category reveals
that the aggregate trend is not uniform: the density gap is driven by
the Open category ($0.190$ vs.\ $0.128$, $213/97$ episodes), whereas Long-Horizon
failures are \emph{denser} than their successes ($0.084$ vs.\
$0.074$, $76/148$). Precision carries the largest failure mass (172 of 443
failures) with almost no density separation ($0.117$ vs.\ $0.112$).
The reversal is not confined to density: Long-Horizon successes run
\emph{staler} than their failures ($14.3$ vs.\ $13.2$), while Memory
reverses the density direction from the other side ($0.108$ vs.\
$0.092$, $79/26$). The practical reading is that scheduling headroom is task-conditional,
which a single fixed ratio cannot exploit and which our protocol is
designed to expose.

\begin{figure}[t]
\centering
\includegraphics[width=\linewidth]{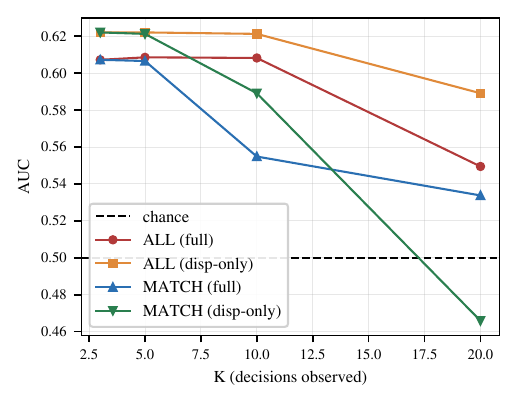}
\caption{Early-warning AUC as a function of prefix length $K$ under
four evaluation settings. The permuted-label null is $0.50$ for every
series; significance vanishes as the observed prefix grows.}
\label{fig:censor}
\end{figure}

\paragraph{Calibration reliability, evaluated against its own leakage.}
A logistic predictor over whole-episode scheduling features reaches
AUC $0.933$ (5-fold CV; permuted-label null $0.50$, $p = 0.010$) for
failure, apparently a strong early-warning capability. It is mostly
not one. Restricting to the first $K$ decisions and controlling
censoring (Fig.~\ref{fig:censor}) gives a
layered picture:

\begin{itemize}
\item \textbf{Post hoc} (whole episode): AUC $0.933$. Dominated by
episode length, which is outcome-dependent by construction.
\item \textbf{Leak-free, early} (first $K \le 10$ decisions, dispatch
statistics only): AUC $0.62$, $p = 0.010$ against the permuted-label
null. Real, but modest, and \emph{not} requiring length or density
features.
\item \textbf{Censoring-matched} (episodes long enough that the prefix
is at most half the trajectory): the signal decays with $K$
($0.62 \to 0.59$ at $K{=}10$, $0.47$ at $K{=}20$, $p = 0.81$).
\end{itemize}

Two conclusions follow. Substantively, early dispatch irregularity
\emph{does} forecast failure, but weakly and only from the first few
decisions: schedule features are worth including in a router and
unsafe to rely on alone, so routing should fuse them with online state
uncertainty, which history-only gates such as Jev do not consult.
Methodologically, the protocol earns its keep: without the leak-free
and censoring controls, the same data would have supported an
early-warning claim roughly $1.5\times$ stronger than is defensible.

\subsection{Decision-Level Routing Frontier}
\label{sec:exp-e32}

The scores above are episode-level summaries. A router, however, acts
at individual decision points: query the slow module, or execute one
more segment of the standing plan (the action Jev's gate approves). We
therefore rescore the release at the decision level, 33{,}443 decision
points over the 929 episodes, with strictly prefix-based online
features (dispatch mean, dispersion, short-horizon fraction, density,
a rejection-rate feature built from request--turn divergence, and the
last dispatch length), under episode-grouped cross-validation so that
no score ever sees its own episode.

\paragraph{Per-decision signal.} The leak-free per-decision AUC is
$0.646$ over the first ten decisions of each episode and $0.636$ over
the first twenty, matching the episode-level $0.62$: the signal
survives the descent to the granularity at which scheduling acts. The
category structure sharpens rather than blurs: Open $0.74$,
Long-Horizon $0.54$, Precision $0.52$, and Memory $0.45$, the last
\emph{below chance} --- the reversal visible in aggregate above, now
visible per decision.

\paragraph{Censoring at the decision level.} The matched-cohort
control behaves as the episode-level analysis predicts: the AUC holds
at $0.649$ (ten-decision prefix) but decays to $0.592$ (twenty). Part
of the longer-prefix signal is censoring-induced, so a deployed router
should be calibrated against the matched numbers --- the endogeneity
correction the protocol requires, here confirmed at the decision
level.

\begin{figure}[t]
\centering
\includegraphics[width=\linewidth]{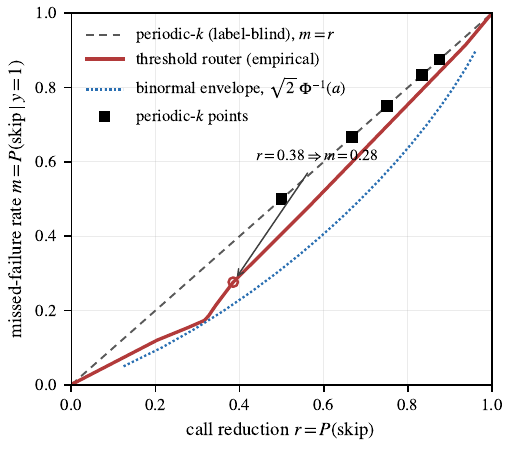}
\caption{Decision-level routing frontier on the 929-episode RoboICL
deployment traces ($33{,}443$ decision points, leak-free
episode-grouped scores, detection rate $a=0.64$).
The threshold router dominates every label-blind period at matched
call reduction; the binormal envelope of
Proposition~\ref{prop:frontier} bounds it from above.}
\label{fig:routing-e32}
\end{figure}

\paragraph{The frontier.} Figure~\ref{fig:routing-e32} reads the
threshold rule through Proposition~\ref{prop:frontier}. The empirical
curve lies strictly below the diagonal everywhere: at the measured
operating point $r = 0.38$ the router misses $m = 0.28$ against the
periodic policy's $m = r = 0.38$, and at $m \le 0.20$ it sustains $r$
up to $0.33$ --- the lower edge of the $33$--$48\%$ reduction band
that the deployed Jev gate reports. The binormal envelope
(Eq.~\eqref{eq:frontier} at $a = 0.64$) sits $5$--$20\%$ above the
empirical curve across the operating range, as the proposition
anticipates for a score that is not a likelihood ratio. A history-only
router with these features therefore reaches the deployed gate's
reduction regime on this release, with its price stated on an
explicit, reusable axis ($m$) rather than as a task-count delta.

\paragraph{Certified dispersion.} The decision count is a misleading
sample size on this release. Because labels are episode-constant and
prefix features overlap within an episode, the router's skip
indicators are strongly clustered: Proposition~\ref{prop:dispersion}
applied to the measured traces gives intraclass correlations
$\kappa = 0.91$--$0.99$ across cohorts and design effects
$\mathrm{DEFF} = 7.6$--$17.6$, so the effective sample size at
twenty decisions per episode is about one thousand, not the nominal
$16{,}449$. The correction is consequential at the headline
operating point: the naive binomial interval at the headline point
($r \approx 0.39$, $m = 0.28$) spans
$\pm 0.007$, while the design-effect certificate spans $\pm 0.030$
and an episode-level block bootstrap spans $\pm 0.029$ --- the
closed-form correction reproduces the resampling gold standard to
within a few percent, on every cohort block. Any deployment claim of
the form ``$r \le 0.33$ at $m \le 0.20$'' should therefore be read
with episode-count precision, not decision-count precision, and the
$33$--$48\%$ reduction band quoted above is consistent with this
certified width.

\paragraph{The high-asymmetry anchor.} The
release also pins the cost side. Its Table~7 reports $9407$ API
seconds over $181$ slow calls (Align Blocks) and $4192$ over $116$
(General Pickup), i.e.\ $36$--$52$ s per slow query, while the fast
system executes at $25$ Hz, $40$ ms per step. The cost asymmetry of
this system family is therefore $\rho \approx 0.9$--$1.3 \times
10^{3}$, three orders of magnitude above the fine-tuned FiS-VLA's
measured $\rho = 0.95$--$1.53$
(Section~\ref{sec:exp-latency}). With measured endpoints at both
tails, the continuum of Section~\ref{sec:continuum} is no longer a
conceptual axis but an instrumented one. At $\rho \approx 1$ sparse
refresh is already Pareto optimal and conditional routing has only
heterogeneity headroom (Section~\ref{sec:exp-prediction}); at $\rho
\approx 10^{3}$ even one-segment reuse pays for itself --- the regime
where the missed-failure rate $m$, not the call cost, becomes the
binding constraint. The Jev gate's task-dependent success drop (Align
Blocks $3/5 \to 2/5$) is exactly this binding constraint realized on
the task axis, and Proposition~\ref{prop:frontier} is the instrument
that prices it before deployment.

\subsection{Cost Instantiation: the Low-Asymmetry End, Measured}
\label{sec:exp-latency}

The router's rule compares $\mathrm{VoI}(s_t)$ against
$c_{\text{slow}}$, and the synthetic testbed holds the query budget
fixed; neither pins down what a query costs in wall-clock terms. We
therefore measure both sides of the asymmetry on the released
FiS-VLA checkpoint (7B), the only one of the three source systems
with public weights. The model runs in bfloat16 on a single RTX 4090
(15.3 of 24\,GB), through the reference dual-system API and deployment
settings: a slow query is one \texttt{slow\_system\_forward} call, a
fast action is one 4-step DDIM \texttt{fast\_system\_forward} call.
Two deviations from the reference defaults were forced by the
checkpoint and are disclosed with the reproducibility notes in
Section~\ref{sec:discussion}: classifier-free guidance is
unavailable, and the fast path needs an in-memory fix to the reference
KV-cache indexing.

\begin{table}[t]
\centering
\footnotesize
\caption{Latency of the FiS-VLA system on one RTX 4090
(reference code path, $10$--$12$ timed runs after warmup; run-to-run
spread $53.8$--$61.1$\,ms for the fast action, $52.0$--$52.1$\,ms for
the slow query). One slow refresh costs about one fast action, or
$3.8$ denoising steps.}
\label{tab:latency}
\begin{tabular}{@{}lcc@{}}
\toprule
Quantity & Wall-clock & Rate \\
\midrule
Slow query (latent refresh) & $52.0$\,ms & $19.2$\,Hz \\
Fast action (4-step DDIM) & $54.8$\,ms & $18.2$\,Hz \\
\quad one denoising step & $13.7$\,ms & $73$\,Hz \\
\midrule
$c_{\text{slow}}/c_{\text{fast}}$, per action & \multicolumn{2}{l}{$0.95$} \\
$c_{\text{slow}}$ in denoising steps & \multicolumn{2}{l}{$3.8$} \\
Fixed $1{:}4$ compute overhead & \multicolumn{2}{l}{$+24\%$} \\
\bottomrule
\end{tabular}
\end{table}

\paragraph{A measurement protocol for $\rho$.} The continuum's
predictions are stated in $\rho$, so its units must be fixed before
numbers from different systems can be compared. We define
\begin{equation}
\label{eq:rho}
\rho \;=\; t_{\text{slow}}\,/\,t_{\text{fast}},
\end{equation}
where $t_{\text{slow}}$ is the wall-clock time of one slow
invocation, transport included, and $t_{\text{fast}}$ the wall-clock
time of one fast control step. Both are end-to-end: $t_{\text{slow}}$
counts whatever the deployed system pays per slow call (network
round trip, API queueing, token generation), not only model forward
time; $t_{\text{fast}}$ is the period at which the fast loop actually
steps. The definition is
operational by design, because scheduling trade-offs are paid in
wall-clock rather than FLOPs: weights open, Eq.~\eqref{eq:rho} is
measured on the deployment path; weights closed, it is read from
reported latencies. Table~\ref{tab:rho} applies it to the five
systems this paper touches. The span it finds, $0.95$ to
${\sim}10^{3}$, is three orders of magnitude, and every prediction
of the continuum inherits this scale.

\begin{table}[t]
\centering
\footnotesize
\setlength{\tabcolsep}{2.5pt}
\caption{The cost-asymmetry continuum under one protocol
(Eq.~\eqref{eq:rho}). ``Meas.''\ rows are this paper's runs
(Table~\ref{tab:latency} for FiS-VLA); ``rep.''\ rows read published
latencies and control rates. The empirical span is three orders of
magnitude, and the LIBERO row shows that co-locating both modules on
one GPU already moves a separate-VLM system an order of magnitude
down the axis.}
\label{tab:rho}
\begin{tabular}{@{}llll@{}}
\toprule
System & Slow call & Fast step & $\rho$ \\
\midrule
FiS-VLA~\cite{fis2025} & $52.0$\,ms local & $54.8$\,ms & $0.95$ (meas.) \\
LIBERO (ours) & $0.46$--$2.2$\,s & $300$\,ms & $1.5$--$7.2$ (meas.) \\
Hume~\cite{hume2025} & VLM at $4$\,Hz & $90$\,Hz ctrl & ${\sim}22$ (rep.) \\
DualVLN~\cite{wei2025dualvln} & VLM consult & fast step & $10$--$20$ (rep.) \\
RoboICL~\cite{roboicl2026} & $36$--$52$\,s API & $40$\,ms chunk & ${\sim}10^{3}$ (rep.) \\
\bottomrule
\end{tabular}
\end{table}

The LIBERO row is a same-device anchor we add to close the gap
between the two measured extremes. The fast module is OpenVLA-7B
executing one control step of LIBERO-Spatial ($299.6$\,ms median
end-to-end on one RTX~4090, bf16), the slow module is
Qwen2.5-VL-7B-Instruct answering a grounded re-plan query on the
rendered observation: $458$\,ms for a one-sentence subgoal
($\rho = 1.53$) and $2.15$\,s for a grounded describe-then-plan
response over $78$ generated tokens ($\rho = 7.19$). Both modules
run on the same card, so neither network transport nor queueing
inflates the ratio; the measurement isolates architecture. Two
lessons follow. First, $\rho$ is a deployment property, not an
architectural one: the same separate-VLM design that costs
$\rho \approx 10^{3}$ behind an API costs $\rho < 10$ when both
weights are local. Second, the cost of a slow call is governed by
its output contract: lengthening the plan from one sentence to a
grounded paragraph multiplies $t_{\text{slow}}$ by $4.7$ at fixed
weights, so the schedule's cost parameter is set jointly by the
model and the response format it is prompted into.

Placed on the continuum of Section~\ref{sec:continuum}, these numbers
carry a prediction, not just a cost. The measured
$c_{\text{slow}}/c_{\text{fast}} = 0.95$ locates FiS at the extreme
low-asymmetry end, opposite the separate-VLM systems ($\rho \approx
10$--$20$ for Hume and DualVLN) that concurrent triggers target. At
$\rho \approx 1$ the periodic baseline of
Proposition~\ref{prop:continuum} is already efficient: the fixed
$1{:}4$ cadence spends $+24\%$ compute, and the residual headroom of
state-adaptive triggers is bounded by collapse heterogeneity alone.
Measured in denoising steps, a refresh is worth $3.8$ fast steps;
this is the currency a latency-budgeted router trades in, and it is a
measured quantity rather than an assumption. The fast action's cost
is dominated by perception: point-cloud preprocessing ($7.4$\,ms), the
vision backbone ($3.3$\,ms) and the projector ($0.1$\,ms) account for
$10.8$ of the $14.0$\,ms that the timed submodules attribute to each
denoising step, while the two cached LLM layers that make the fast
path fast contribute $3.2$\,ms. The fast-only control rate we
measure, $18.2$\,Hz, is consistent with the reported $21.9$\,Hz up to
hardware.

This grounds the cost side only. Whether calibrated routing shifts
FiS's frontier at equal task success is the integration half of the
roadmap (Section~\ref{sec:roadmap}), and by
Proposition~\ref{prop:continuum} its outcome is a testable
prediction: any frontier shift at $\rho \approx 1$ must come from
collapse heterogeneity, so a null or small result would calibrate the
continuum rather than falsify the router.

\subsection{Testing the Continuum Prediction on the Testbed}
\label{sec:exp-prediction}

Proposition~\ref{prop:continuum} makes a falsifiable claim at the
measured operating point: with $\rho = 0.95$ fixed, the advantage of
calibrated routing over the fixed $1{:}4$ schedule should be a
function of collapse \emph{heterogeneity} alone, vanishing in the
homogeneous regime. We test this directly. Holding the time-averaged
collapse rate fixed at the E0-v2 level, we vary the fraction $w$ of
hard steps from $0$ (homogeneous collapse, the regime the proposition
says makes periodic schedules optimal) to $0.25$, and re-run the
budget-matched comparison of Section~\ref{sec:exp-noise} at each
level. All budgets translate into wall-clock overhead through the
measured $\rho = 0.95$: the fixed $1{:}4$ baseline spends $q=0.250$,
a $+23.7\%$ compute overhead, matching the measured $+24\%$ of the
deployed cadence.

\begin{table}[t]
\centering
\footnotesize
\caption{Budget-matched advantage of calibrated routing over the
fixed $1{:}4$ schedule as collapse heterogeneity grows
(mean-matched; 5 seeds, 100 episodes of 120 steps each). The
advantage vanishes in the homogeneous regime and grows monotonically,
as Proposition~\ref{prop:continuum} predicts.}
\label{tab:prediction}
\begin{tabular}{@{}lcccc@{}}
\toprule
Heterogeneity $w$ & 0 & 0.05 & 0.10 & 0.25 \\
\midrule
Advantage $\Delta$ & $+0.005$ & $+0.019$ & $+0.022$ & $+0.059$ \\
$95\%$ CI & $\pm.004$ & $\pm.005$ & $\pm.004$ & $\pm.005$ \\
\bottomrule
\end{tabular}
\end{table}

The prediction holds (Table~\ref{tab:prediction}). At $w=0$ the
advantage is $+0.005 \pm 0.004$, statistically indistinguishable from
the zero the proposition predicts; the small residual traces to the
budget-matching artefact noted in Section~\ref{sec:exp-noise} (the
closest threshold realises $q = 0.269$, a $7.6\%$ overspend). As $w$
grows the advantage rises monotonically to $+0.059 \pm 0.005$ at the
E0-v2 heterogeneity level. The low-$\rho$ content of the continuum is
therefore not an assumption but a measured regularity: at
$\rho \approx 1$, \emph{all} of the headroom an adaptive trigger can
exploit is the heterogeneity term, and a homogeneous-collapse
deployment cannot beat a well-tuned fixed ratio at any threshold.
What remains is to test the transfer of this prediction to the
deployed FiS checkpoint itself, which the next section does.

\subsection{The Measured Frontier on the Fine-Tuned Checkpoint}
\label{sec:exp-frontier}

The backbone-level integration was previously blocked by missing
fine-tuned weights. We remove the blocker by producing them: the
open 7B checkpoint is fine-tuned on four RTX 4090s (96\,GB total)
with fully sharded data parallel training, head-only (the action head,
input, time, and final layers; $139.7$M of $7.6$B parameters, the
largest subset that the memory budget permits), for 740 steps over two
epochs on 320 RLBench episodes from the four reference tasks
(episodes $0$--$79$ per task, generated through the reference demo
pipeline); the training loss falls from $20.2$ to $0.42$. Three
disclosures attach to this setup. First, head-only training caps the
policy's competence, which matters below. Second, the
pretrained checkpoint ships OXE statistics only, so RLBench
fine-tuning is mandatory and the normalisation statistics are
recomputed from our data. Third, the reference inference path
unconditionally requires point-cloud inputs while the reference
training path with \texttt{load\_pointcloud=false} inserts none; we
patch the inference path to match training.

On this checkpoint we measure, on episodes $80$--$89$ per task
(disjoint from training), the three quantities the scheduling theory
consumes.

\begin{figure*}[t]
\centering
\includegraphics[width=\textwidth]{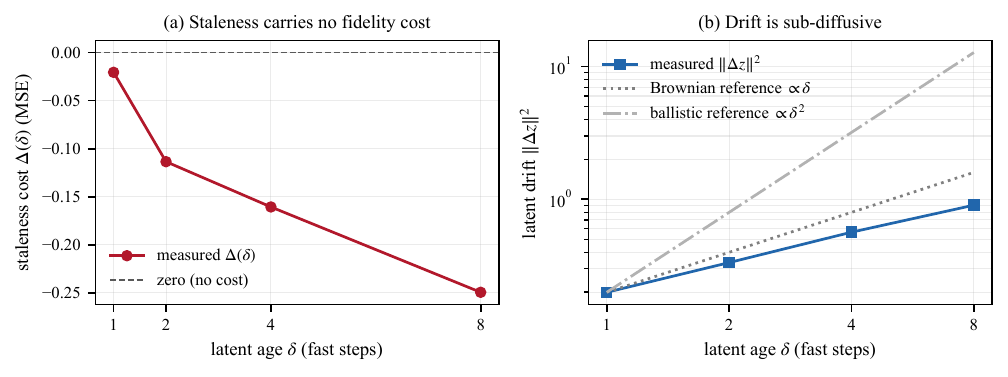}
\caption{Staleness and drift on the fine-tuned checkpoint, held-out
episodes. (a) The staleness-cost curve $\Delta(\delta)$ is
nonpositive: an aged slow latent costs nothing. (b) The latent
itself drifts sub-diffusively, linearly in age rather than as a
random walk.}
\label{fig:staleness}
\end{figure*}

\emph{(i) The staleness-cost curve.} For latent ages
$\delta \in \{1,2,4,8\}$ fast steps, we run the fast system on each
held-out frame once with a fresh slow latent and once with a latent
aged $\delta$ steps, and record
$\Delta(\delta) = \mathrm{MSE}_{\text{aged}} -
\mathrm{MSE}_{\text{fresh}}$
(Figure~\ref{fig:staleness}a). Staleness carries no fidelity cost
anywhere in the range:
$\Delta = (-0.02, -0.11, -0.16, -0.25)$ for
$\delta = (1, 2, 4, 8)$; the trend is mildly negative, not positive.
The curve is also task-conditional in a structured way:
\texttt{close\_jar} and \texttt{sweep} are negative at every age,
\texttt{open\_drawer} turns negative after $\delta{=}1$, and
\texttt{place\_wine} is the only task with a positive cost at
$\delta{=}1$ ($+0.057$), consistent with its dependence on visually
tracking a target that the arm is carrying; by $\delta{=}4$ even
this task is at or below zero. Task-conditionality at the
value-function level echoes the category-dependence of the trace
analysis (Section~\ref{sec:exp-roboicl}).

\emph{(ii) Latent drift.} The aged latents themselves do move:
$\|\Delta z\|^{2} = (0.20, 0.33, 0.57, 0.90)$ over the same ages,
growing linearly in $\delta$ ($R^{2} = 0.91$) rather than
quadratically ($R^{2} = 0.24$): the RMS displacement follows the
$\sqrt{\delta}$ scaling of a Wiener process, with a log-log slope of
$0.72$, mildly sub-diffusive (Figure~\ref{fig:staleness}b). The
displacement is real, but the action map is locally insensitive to
it at this scale, which is precisely what the flat staleness cost
asserts.

\emph{(iii) The influence control.} Flat staleness cost would be
degenerate if the fast system ignored the latent. It does not:
replacing the latent with one computed from a different task's
episode raises action MSE by $+0.275$ ($+11\%$), an order of
magnitude more than any staleness effect. The latent carries
task-relevant information, and that information simply does not
decay over the measured horizon.

\begin{figure*}[t]
\centering
\includegraphics[width=\textwidth]{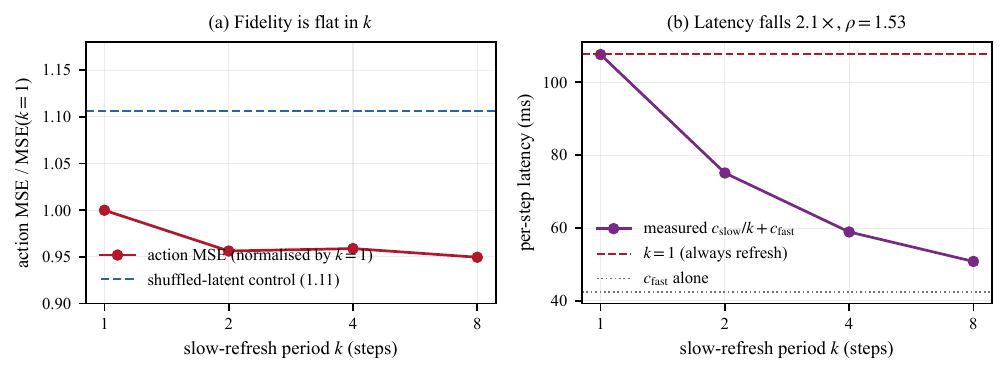}
\caption{The measured frontier. (a) Action MSE is flat in the
refresh period $k$; the shuffled-latent control ($+11\%$) marks the
scale at which the latent does matter, an order of magnitude above
any staleness effect. (b) Amortised per-step latency falls
$2.1\times$ over the same range.}
\label{fig:measured-frontier}
\end{figure*}

The frontier follows
(Figure~\ref{fig:measured-frontier}, Table~\ref{tab:frontier}). Under bf16
inference on one 4090 the measured costs are
$c_{\text{slow}} = 65.3$\,ms and $c_{\text{fast}} = 42.5$\,ms, so
$\rho = 1.53$ (the fp32 release path of
Section~\ref{sec:exp-latency} gives $0.95$; precision shifts
$\rho$, but both values sit in the same low-asymmetry regime).
Amortised per-step latency falls from $107.6$\,ms at $k{=}1$ to
$50.8$\,ms at $k{=}8$, a $2.1\times$ reduction at equal or better
action fidelity. With $\Delta(\delta) \le 0$ over the measured
range, the renewal-reward average cost
$C(k) = c_{\text{slow}}/k + c_{\text{fast}} +
\frac{1}{k}\sum_{\delta \le k}\Delta(\delta)$ is strictly decreasing
in $k$: the periodic baseline of Proposition~\ref{prop:continuum}
has no interior optimum here, and the drift instance of
Remark~\ref{rem:drift} is the applicable model, with a drift
intensity so small that the predicted optimal period exceeds the
measured range. In-distribution, refreshing the slow latent more
often than once every eight steps buys nothing and costs up to
$2.1\times$ latency; the efficient frontier point is the sparsest
one measured.

\begin{table}[t]
\centering
\footnotesize
\caption{The measured frontier of the fine-tuned FiS-VLA on held-out
RLBench demonstrations (teacher-forced, 16 episodes from 4 tasks,
bf16, one RTX 4090; latency amortised as
$c_{\text{slow}}/k + c_{\text{fast}}$ with
$c_{\text{slow}} = 65.3$\,ms, $c_{\text{fast}} = 42.5$\,ms
measured in the same runs). Fidelity is flat in $k$ while latency
falls $2.1\times$.}
\label{tab:frontier}
\begin{tabular}{@{}lcccc@{}}
\toprule
Refresh period $k$ & 1 & 2 & 4 & 8 \\
\midrule
Action MSE & $2.431$ & $2.325$ & $2.331$ & $2.308$ \\
Per-step latency (ms) & $107.6$ & $75.1$ & $58.9$ & $50.8$ \\
\bottomrule
\end{tabular}
\end{table}

\emph{Closed-loop status.} We also attempted closed-loop RLBench
evaluation at $k \in \{1, 4\}$, twice: with the head-only policy
and, after it failed, with a LoRA fine-tune ($r{=}16$ on the LLM
attention and MLP projections, $179.6$M trainable parameters
including the head) under the same data and budget, loss $0.4404$.
Neither policy succeeded in any of the $20$-episode runs per task
and period, and both failed the same way: episodes self-terminate
within $1$--$23$ steps when the motion planner rejects a commanded
end-effector target as unreachable. The mechanism is competence,
not staleness. The fine-tuning data are keyframe trajectories whose
per-step displacements reach half a metre; the fine-tuned policy
reproduces them with centimetre-level open-loop error on held-out
frames, and that error compounds through the proprioceptive
feedback loop until a commanded pose leaves the workspace, ending
the episode. Three controls support this reading. Correcting a
normalization defect in the reference code path, disclosed in
Section~\ref{sec:discussion}, changes neither the open-loop error
nor the closed-loop outcome. The failure is a property of the
action head rather than of the latent's age, so no refresh period
can repair it; the shuffled-latent control above bounds how much
any latent content could matter. And the terminated episodes are
too short for the slow latent's age to vary meaningfully across
schedules. This is the pre-registered risk of a short fine-tuning
budget, realized twice under two fine-tuning regimes; by
Remark~\ref{rem:attribution} the competence floor saturates the
metric, so the comparison is uninformative for scheduling rather
than a null result: it cannot detect the freshness value even if
that value were positive. A success-level frontier comparison
requires the full fine-tune that a 96\,GB budget cannot fit
(Section~\ref{sec:roadmap}).

\subsection{Compiling the Measured Value Function into Schedules}
\label{sec:exp-distill}

The measurements so far are diagnostic; the compiler turns them into
schedules. The distillation loop runs in three stages. \emph{Measure}
probes a task's staleness value function $\Delta(\delta)$ on held-out
episodes; \emph{compile} evaluates the renewal cost
$C(k) = c_{\text{slow}}/k + c_{\text{fast}} +
w_{\text{fail}}\,\overline{\Delta}(k)$ on a grid of periods and
returns the minimizer $k^{*}$; \emph{validate} rolls closed-loop
periodic-$k$ episodes and compares the compiled period with the
oracle period $k^{**}$ that minimizes the empirical closed-loop
cost. One convention matters enough to state up front: the
validation loop must share the compiler's accounting. The slow
module's benefit is carried entirely by the latent refresh inside
$\Delta$, with no separate rescue term, and the latent age resets
only on queries; a harness that rescues failed steps unconditionally
makes every query look worthless and pushes every oracle period to
the boundary. We hit exactly this artefact, fixed it before taking
any number, and flag it because it is the loop's main pitfall, not
because it is ours.

Two cases bracket the loop. On the real checkpoint
(Section~\ref{sec:exp-frontier}) the measured $\Delta(\delta)$ is
nonpositive for every task at every age probed, so the compiler sends
all four tasks to the boundary of the probed range: the correct
decision when freshness is free, and one that requires no learning,
only the measured curve. To exercise the interior, the same compiler
is fed controlled task families whose $\Delta(\delta)$ is synthetic
but whose $C(k)$ is validated against closed-loop rollouts
(Table~\ref{tab:distill}). Across families and probe budgets from
$200$ to $20{,}000$ probe steps per age, the compiled period tracks
the oracle at a cost gap of at most $2.4\%$; at the largest budget it
recovers the interior optimum exactly where the optimum is interior
and stays at the boundary where staleness is flat. Measurement also
buys schedulable value quickly and with diminishing returns: across
families the worst cost gap falls from $7.8\%$ at $200$ probe steps
per age to $2.0\%$ at $20{,}000$, and the staleness-free family
needs no probes at all. Scheduling knowledge is therefore cheap in
probe steps even where it is expensive in rollouts. The point is not
that the compiler wins on this testbed. It is that the
measure--compile--validate loop closes at all, and that its decisions
are driven by the sign and shape of the measured value function: the
real system lands on the boundary because its measured $\Delta$ says
so, the synthetic families land in the interior for the same reason.
Adaptive triggers, periodic baselines, and compiled schedules thereby
become commensurable as different amounts of measurement invested in
the same $C(k)$, and the compiler is the limiting case: measure
once, schedule without further adaptation, accept the cost of stale
information between measurements.

\paragraph{Compilation as sample-average approximation.}
The compile step is an instance of sample-average approximation: it
minimizes a cost estimated from $N$ Bernoulli probes. The
identification brings machinery that ad hoc validation lacks: a
statistical upper bound on the optimal cost (re-evaluating the
compiled period on an independent large sample) and a lower bound
(averaging batch optima over independent probe
batches~\cite{mak1999}), whose difference certifies the compiled
schedule's optimality gap with a confidence
interval~\cite{homemdemello2014}. Against the synthetic families'
known ground truth the certificates are valid in $143$ of $144$
instances and conservative at small budgets; the certified gap
decays close to the $N^{-1/2}$ rate (per-decade ratios $2.6$--$4.7$
against the $3.2$ ideal), reaching $0.002$--$0.004$ cost units at
$N=20{,}000$. The worst realized gaps of Table~\ref{tab:distill}
($7.8\%$ at $N=200$, $2.0\%$ at $N=20{,}000$) sit inside the
certified envelopes, and their shrinkage, a $100\times$ probe
investment for a $4\times$ gap reduction, is the flat-bowl accounting
of Remark~\ref{rem:eoq}: a two-fold period error costs only
$1.04$--$1.11$ against the $1.25$ identity value, flattest for the
sublinear drift family.

\paragraph{Allocating the validation budget.} The certificates say
how good a compiled schedule is at a given budget; the allocation
rule of Remark~\ref{rem:ocba} says where the next probe should go.
On the same four families ($20{,}000$ rollouts per candidate period,
$K = 16$), costs are homoscedastic across periods, so the OCBA ratio
Eq.~\eqref{eq:ocba} reduces to gap-proportional allocation
$n_k \propto \Delta_k^{-2}$, tested against uniform allocation. At
$N = 2{,}000$ validation episodes, gap-proportional allocation raises
the probability of recovering the oracle period from $0.33$--$0.61$
to $0.67$--$0.93$, with the near-tie cluster (suboptimality below one
cost unit) absorbing $95$--$98\%$ of the budget; at $N = 200$ no
allocation separates the optimum, the floor being set by the bowl's
width rather than the allocator, the same saturation the EOQ identity
predicts. Adaptive re-planning tracks the static allocation at small
budgets and gives back a little at intermediate ones, which is the
practical content of the remark: the $\sigma/\Delta$ structure
carries the selection accuracy, and it can be compiled once from a
pilot measurement instead of re-estimated online.

\begin{table}[t]
\centering
\footnotesize
\caption{Schedule compilation from the measured value function.
Synthetic families ($60$-step episodes, $400$ validation episodes
per period, $5$ seeds): $k^{**}$ minimizes the empirical closed-loop
cost, $k^{*}$ is compiled from probed $\Delta(\delta)$ at $20{,}000$
probe steps per age, cost gap is
$|C(k^{*}) - C(k^{**})|/C(k^{**})$ (worst over seeds). The real FiS
row repeats the measured result of Section~\ref{sec:exp-frontier}:
nonpositive $\Delta$ compiles every task to the edge of the probed
range.}
\label{tab:distill}
\begin{tabular}{@{}llccl@{}}
\toprule
Family & $\Delta(\delta)$ shape & $k^{**}$ & $k^{*}$ & cost gap \\
\midrule
tracking & $+b_0\,\delta$ & 5 & 5 & $0.0\%$ \\
drift & $+b_0\sqrt{\delta}$ & 3 & 5 & $2.0\%$ \\
contact & $-c_0$, $\delta \ge 1$ & 15 & 15--16 & $1.9\%$ \\
flat & $0$ & 16 & 16 & $0.0\%$ \\
\midrule
FiS (4 tasks, meas.) & $\le 0$, $\delta \le 8$ & --- & 8 (edge) & --- \\
\bottomrule
\end{tabular}
\end{table}

\subsection{A Closed-Loop Test at the Middle of the Continuum}
\label{sec:exp-libero}

The continuum is anchored at its two ends by measured systems
(Table~\ref{tab:rho}); the middle is where the separate-VLM
literature's economic argument is usually assumed to hold. We
instantiate the protocol there on a real dual system. The fast module
is OpenVLA-7B fine-tuned on LIBERO-Spatial ($6000$ steps, LoRA
$r{=}16$; $57\%$ mean success over the $10$ tasks under the official
evaluation conventions, including the unnormalisation key, the
$180^{\circ}$ image convention and the gripper transform of the
reference pipeline). The slow module is Qwen2.5-VL-7B-Instruct
answering a grounded query on the rendered observation: $458$\,ms for
a one-sentence subgoal against the fast module's $299.6$\,ms control
step, so $\rho = 1.53$ (Section~\ref{sec:exp-latency}). Both modules
run on one RTX~4090, which places the operating point between FiS
($\rho = 0.95$) and the separate-VLM systems
($\rho \approx 10$--$20$), at the regime the continuum says is
contested.

The slow module's action channel is a subgoal rewrite: its
one-sentence reply replaces the instruction the fast policy
conditions on until the next query. Eleven arms vary the channel, its
dose and its undo at a matched slow budget. Arm~A queries nothing. Arms~B
and~B2 query every $25$ steps; B2 additionally cleans the reply into
an imperative, because the raw reply often begins with a meta prefix
(``the next subgoal is to\,\dots''). Arms~C and~C2 query when the
end-effector displacement over a $15$-step window falls below
$2$\,cm, the stall trigger of Remark~\ref{rem:sprt}; C2 applies the
same cleaning. Arm~D screens the rewrite: a cheap binary progress
verdict is sampled every $25$ steps and the expensive rewrite fires
only on a negative verdict. Arm~E3 replaces the language channel
entirely: on the same stall trigger, the slow module ranks $k{=}3$
sampled candidate fast actions by their imagined $h{=}6$-step
rollouts and the selected candidate's first action is executed, so
the slow module never touches the instruction. The extension adds
six arms that close the two gaps this family leaves. Arms~k50 and
k10 rerun the cleaned rewrite at periods $50$ and $10$, halving and
more than doubling the adaptive arm's rate, so the channel's
dose-response is measured rather than assumed; arms~e005 and~e001
move the ranking trigger to $\varepsilon = 0.05$ and
$\varepsilon = 0.01$ for the same reason on that channel. Arm~E3b
reruns the ranking channel with the candidate
order shuffled in front of the slow module, separating ranking
ability from presentation order. Arm~E4 attaches an undo to the
rewrite: on a stall that arrives while a rewrite is active, the
instruction reverts to the validated task instruction at no query
cost, which screens the intervention by its consequence rather than
the state. Each arm runs $10$ tasks $\times$ $3$ episodes
$\times$ $300$ steps with shared seeds and initial states, and the
table reports the same arms extended to $10$ episodes per task.
Read through Proposition~\ref{prop:channel}, the design separates the
two layers: arms~A--C2 share one channel, a subgoal rewrite whose
transfer coefficient $\eta_{\mathcal{C}}$ is unknown a priori, and
vary only the detection layer, while the arms that follow vary the
channel, its dose, or its undo at a fixed trigger and so price
$\eta_{\mathcal{C}}$ directly.

\begin{table*}[t]
\centering
\footnotesize
\setlength{\tabcolsep}{3.5pt}
\caption{LIBERO-Spatial closed-loop arms at $\rho = 1.53$
(success over $30$ episodes per arm with exact $95\%$
Clopper--Pearson intervals; the ten-episode extension, superscript
$10$, reports the nine tasks common to all arms --- ninety
paired episodes per arm, see the statistical-reading paragraph).
``Overhead'' is slow-query wall-clock over measured
fast-policy wall-clock; the revert channel's undos are local and
consume no query, so its Calls/ep counts rewrites only. The
extension sweeps the rewrite channel over periods
$k \in \{10, 25, 50\}$ and a verification screen, the ranking
channel over trigger thresholds with a shuffled-selection control
that removes the position bias, and adds a revert channel that
undoes a rewrite that fails to restore progress. The bottom block
replicates arms~A, B2 and~E4 on LIBERO-Goal, a second task suite
fine-tuned and evaluated under the same protocol with an LLM slow
endpoint (mean call latency $600$\,ms); its overhead is slow-call
wall-clock over the fast-only arm's wall-clock on the same tasks.}
\label{tab:libero-arms}
\begin{tabular}{@{}lllcccc@{}}
\toprule
Arm & Channel & Trigger & Success & $95\%$ CI & Calls/ep & Overhead \\
\midrule
A  & none            & ---            & $17/30$ & $[.37,.75]$ & $0$   & $0\%$ \\
B  & rewrite, raw    & periodic $25$  & $8/30$  & $[.12,.46]$ & $10.4$ & $8.9\%$ \\
B2 & rewrite, clean  & periodic $25$  & $11/30$ & $[.20,.56]$ & $9.4$  & $10.8\%$ \\
C  & rewrite, raw    & stall          & $13/30$ & $[.26,.63]$ & $10.7$ & $9.0\%$ \\
C2 & rewrite, clean  & stall          & $13/30$ & $[.26,.63]$ & $10.7$ & $13.0\%$ \\
\midrule
A$^{10}$  & none            & ---            & $52/90$ & $[.47,.68]$ & $0$    & $0\%$ \\
\multicolumn{7}{@{}l}{\emph{rewrite family: dose and screening sweep}}\\
B2$^{10}$ & rewrite, clean  & periodic $25$  & $39/90$ & $[.33,.54]$ & $9.1$  & $7.0\%$ \\
C2$^{10}$ & rewrite, clean  & stall          & $43/90$ & $[.37,.59]$ & $10.3$ & $7.8\%$ \\
D$^{10}$  & rewrite, screened & verify $25$  & $43/90$ & $[.37,.59]$ & $16.7$ & $10.7\%$ \\
k50$^{10}$& rewrite, clean  & periodic $50$  & $36/90$ & $[.30,.51]$ & $4.8$  & $5.4\%$ \\
k10$^{10}$& rewrite, clean  & periodic $10$  & $34/90$ & $[.28,.49]$ & $23.8$ & $27.1\%$ \\
\multicolumn{7}{@{}l}{\emph{ranking family: trigger sweep and shuffled control}}\\
E3$^{10}$ & action critic   & stall          & $56/90$ & $[.51,.72]$ & $6.4$  & $1.8\%$ \\
E3b$^{10}$& ranking, shuffled & stall        & $57/90$ & $[.53,.73]$ & $6.6$  & $2.0\%$ \\
e005$^{10}$& ranking, shuffled & stall $.05$ & $54/90$ & $[.49,.70]$ & $9.5$  & $2.8\%$ \\
e001$^{10}$& ranking, shuffled & stall $.01$ & $57/90$ & $[.53,.73]$ & $5.1$  & $1.5\%$ \\
\multicolumn{7}{@{}l}{\emph{revert family: intervene, observe, undo}}\\
E4$^{10}$ & rewrite, revert & stall          & $54/90$ & $[.49,.70]$ & $8.0$  & $9.1\%$ \\
\midrule
\multicolumn{7}{@{}l}{\emph{cross-suite replication on LIBERO-Goal (same protocol, second task suite, LLM slow endpoint)}}\\
A$_g$  & none            & ---            & $48/100$ & $[.38,.58]$ & $0$    & $0\%$ \\
B2$_g$ & rewrite, clean  & periodic $25$  & $6/100$  & $[.02,.13]$ & $11.6$ & $10.1\%$ \\
E4$_g$ & rewrite, revert & stall          & $48/100$ & $[.38,.58]$ & $13.1$ & $11.5\%$ \\
\bottomrule
\end{tabular}
\end{table*}

Three results survive the audit of Remark~\ref{rem:exact}.

\paragraph{The rewrite channel is harmful in aggregate and
task-signed in detail.} Every closed-loop arm that rewrites scores
below the fast-only arm (Table~\ref{tab:libero-arms}), but at ten
episodes per task the sign of the channel effect is task-level.
Task~$3$ is the destructive instance: the fast policy solves $8/10$;
the periodic arm rewrites about ten times per episode and solves
$2/10$; the adaptive arm, whose detector correctly sees episodes
stall ($159$ queries on this task), still solves only $4/10$. A
query that returns a plausible-but-wrong subgoal is worse than no
query on the tasks the fast policy had partially solved. The same
channel is \emph{helpful} elsewhere: on task~$0$ both rewrite arms
score $7/10$ against the fast policy's $5/10$, and on task~$1$ the
periodic arm scores $7/10$ against $5/10$; precision-limited
pick-and-place episodes are exactly where a fresh subgoal lands.
The aggregate contrast mixes both directions ($24$ and $11$
discordant pairs for fast-only against the periodic arm) and nearly
cancels; a single success-rate comparison would have read this
system as ``no effect'' when its true state is two opposite,
task-localized effects. This is the channel-mismatch regime of
Remark~\ref{rem:attribution} in closed-loop form, and it is why the
protocol's readout is per-task.

\paragraph{At matched budget, the adaptive trigger allocates queries
the way the threshold rule predicts, but the allocation does not
convert.} The stall arm's per-task budget spans $41$ to $205$ calls
against the periodic arm's $70$ to $108$, and the two arms allocate
differently in kind. The periodic arm's budget tracks episode length
almost perfectly (per-task calls correlate with its own per-task
success at $r = -0.965$): a label-blind schedule spends in proportion
to how long each episode runs, which is the mechanical signature of
blindness, not of attention. The stall arm deviates from that baseline
($r = -0.773$) by reading state: it spends $55\%$ of its queries on
the three tasks with the lowest success rates, against $38\%$ for the
periodic arm, and its two lightest budgets ($41$ and $56$ calls) fall
on tasks~$9$ and~$7$, where the fast policy still solves half or more
of its episodes and where the periodic arm places its heaviest
single-task budget ($108$ calls on task~$9$). The detection layer of
Proposition~\ref{prop:sampling} works; the outcome layer does not. At
ten episodes the adaptive and periodic arms are statistically
indistinguishable ($19$ vs.\ $15$ discordant pairs, exact McNemar
$p = 0.61$); the three-episode hint of an adaptive advantage did not
replicate. This is Proposition~\ref{prop:channel} instantiated from
the other side: when the transfer coefficient $\eta_{\mathcal{C}}$ of
the rewrite channel is low, no detection layer, however good,
converts better query placement into more solved episodes, and the
allocation gain is absorbed by the channel. (At three episodes the
same cleaning that recovers $10$ points on the periodic arm leaves
the adaptive arm unchanged, $3$ of $6$ discordant pairs, which
localizes the raw prefix artefact to the periodic channel, where
every query lands on a freshly written instruction.)

\paragraph{The continuum's directional prediction does not transfer
to this channel.} The continuum predicts that adaptive triggers gain
over fixed ratios as $\rho$ grows, because query economy comes to
dominate; the LIBERO arms sit at a $\rho$ where the adaptive gain
over periodic is already visible in allocation, yet no arm beats
fast-only, so the \emph{sign} of the channel effect, not its
magnitude under the cost model, decides the outcome. We report the
negative rather than reinterpret it: at mid-continuum $\rho$ the
deployment-level VoI of a slow module is a property of the
module--channel--cost triple, not of the slow module's knowledge
(Proposition~\ref{prop:channel}), and
a trigger paper that reports only aggregate success at a single
operating point would read this system as ``the slow module does not
help''. The four-measure protocol instead localizes the failure per
task (rewrite corruption on solved tasks, helpless queries on stuck
ones), which is the actionable statement.

\paragraph{Replacing the channel does not convert either.} Arm~E3
removes the language bottleneck and lets the slow module act directly:
on the stall trigger it ranks three sampled candidate actions by their
imagined six-step outcomes and executes the selected candidate's first
action. The arm scores $56/90$ against fast-only's $52/90$ ($7$ vs.\
$3$ discordant pairs, exact McNemar $p = 0.34$), at $6.4$ stall-gated
calls per episode and $1.8\%$ slow-query wall-clock overhead. The
ranking channel is not free of its own pathologies: each query costs
$18$ additional fast-policy inferences for the imagined rollouts
($14.7\%$ total wall-clock), and the ranker concentrates $524$ of its
$573$ selections on the third candidate, a position bias that
confounds ``the slow module chose well'' with ``the slow module chose
late''. We read the arm as the second point of a channel sweep rather than as
a positive result: at this operating point the rewrite channel is
harmful in aggregate and task-signed in detail
(Proposition~\ref{prop:channel} with $\eta_{\mathcal{C}} < 0$), and
the ranking channel is statistically indistinguishable from no channel
at all ($\eta_{\mathcal{C}} \approx 0$). Neither channel converts the
detection layer's correct query placement into solved episodes, which
is the mid-continuum statement the section exists to make: the
deployment-level value of a slow module is a property of the
module--channel--cost triple, and the two channels the slow module can
express without screening fail their conversion step at
$\rho = 1.53$. Two gaps remain open at this point: the ranking arm's
position bias confounds choice quality with choice position, and the
language channel has never been tested under an actuator that can
undo it. The next two arms close both gaps.

\paragraph{Shuffled selection scores the same as ranked selection.}
Arm~E3b reruns the ranking channel with the candidate order
randomized in front of the slow module, which removes the position
bias while leaving trigger, sampler and budget untouched. The
selections spread from the critic's $524$-of-$573$ concentration on
the third candidate to an even $\{211, 189, 198\}$ split over the
three positions, and the arm scores $57/90$ against the critic's
$56/90$ ($6$ vs.\ $5$ discordant pairs, $p = 1.00$). The slow
module's judgement therefore contributes nothing a coin flip does
not: whatever the ranking channel is worth over fast-only, it is
worth through the \emph{candidates} the sampler puts on the table,
not through the \emph{choice} among them. In the factorized form of
Proposition~\ref{prop:channel}, the sampler moves the distribution
of the correction $\Delta$ that the interface can execute, while the
selector decides which correction is executed; the experiment says
the executed correction's identity is nearly irrelevant and its
mere availability is not. This is the fine structure of the selection
regime of Remark~\ref{rem:channel-risk}: the downside is bounded by
the candidate support, and the upside is bounded by the sampler, so a
channel whose chooser can be replaced by a coin is a channel whose
value the chooser does not price.

\paragraph{The screening channel inherits the channel it screens.}
Arm~D gates the rewrite through a binary progress verdict sampled
every $25$ steps, releasing the expensive rewrite only on a negative
verdict. The verdict is biased: $89\%$ of the $798$ checks answer
``no progress'', so the gate releases nearly every rewrite and the
arm spends $16.7$ slow calls per episode, twice the adaptive arm's
budget. The outcome is the adaptive arm's profile reproduced:
$43/90$, statistically identical to C2 ($16$ vs.\ $16$ discordant
pairs, exact McNemar $p = 1.00$) and indistinguishable from
fast-only ($21$ vs.\ $12$, $p = 0.16$), with the task-signed
structure intact, gains on tasks~$0$, $1$ and~$7$ ($9/10$ on the
latter, the best single-task score in the table), collapse on
task~$3$ ($2/10$). The reason is structural: the verifier screens
the \emph{state} (``is progress stalled?''), not the
\emph{intervention} (``would this rewrite help?''), so it selects
\emph{when} to rewrite with near-perfect recall of stalls and
\emph{whether the rewrite is worth applying} not at all. Screening
adds a second detection layer on top of an already-working one and
leaves the transfer coefficient $\eta_{\mathcal{C}}$ untouched,
which is the third instance of the section's statement: none of the
three channels this slow module can express, raw rewrite, screened
rewrite, or action ranking, converts the detection layer's correct
query placement into solved episodes at this operating point. What
the screening experiment cannot settle is whether the failure is a
property of the channel or of the budget it was given, and the dose
sweep that follows prices exactly that.

\paragraph{The dose axis has no slope, and every dose is negative.}
A note on what this axis is: in simulation the episode horizon is
counted in steps, so an injected latency would not change the success
rate and the delay component of $\rho$ is not identifiable at the
success layer; the sweep therefore moves the query frequency, which is
the executable form of the success-versus-$k$ question, while the
delay component of $\rho$ is anchored by the hardware measurement of
Table~\ref{tab:rho}. The rewrite family is re-run at periods $k = 50$ and $k = 10$,
halving and more than doubling the adaptive arm's query rate, and
the ranking family at trigger threshold $\varepsilon = 0.05$ against
the $0.02$ already measured. The result is flat where a dose-response
should live: $36/90$ at $k = 50$, $39/90$ at $k = 25$ and $34/90$ at
$k = 10$ differ from each other by less than the protocol's
resolution ($p \geq 0.36$ pairwise), while all three sit far below
fast-only's $52/90$, the two extremes at nominal significance
($12$ vs.\ $28$ and $13$ vs.\ $31$ discordant pairs, $p = 0.017$ and
$p = 0.010$). The ranking family is flat in the other direction:
$54/90$ at $\varepsilon = 0.05$ against $57/90$ at $\varepsilon = 0.02$
($8$ vs.\ $11$, $p = 0.65$) and $57/90$ again at
$\varepsilon = 0.01$ ($7$ vs.\ $7$ against the shuffled $0.02$ arm,
$p = 1.0$), so the trigger axis spans a $1.9\times$ query-rate range
($456$ to $851$ slow calls) inside a three-episode band with every
contrast null. The
pre-registered prediction of a negative slope on the rewrite axis is
therefore rejected, and rejected in favour of a stronger statement:
the language channel's failure has no dose at which it succeeds.
Neither querying rarely nor querying often rescues it, because the
loss is not paid at the detection layer the dose axis moves; it is
paid at the transfer coefficient of Eq.~\eqref{eq:channel}, which
scheduling cannot reach. This is the operational content of
Proposition~\ref{prop:channel}(iii) for a practitioner: tuning the
trigger on a channel with $\eta_{\mathcal{C}} < 0$ optimizes the rate
at which value is destroyed. The family-level contrast says the same
thing additively: $43.3\%$ pooled over the five rewrite arms against
$62.2\%$ pooled over the four ranking arms, with every
within-family parameter contrast null, so the channel decision
dominates every scheduling decision the protocol can express.

\paragraph{An actuator that can undo its intervention is the one
positive contrast in the table.} Arm~E4 closes the gap the ranking
arm left: the rewrite fires on the stall trigger as in the adaptive
arm, but if the policy stalls again while a rewrite is active, the
instruction reverts to the validated task instruction, spending no
query. The mechanism runs as a strict alternate loop, $848$ rewrites
answered by $840$ reverts over the ten tasks, almost every rewrite
followed by its own undo at the next stall: the slow module's
subgoals rarely survive contact with the policy for more than a few
steps, and the revert is what keeps them from doing damage. The arm
scores $54/90$, indistinguishable from fast-only ($6$ vs.\ $4$,
$p = 0.75$) with a total downside of a single episode against the
rewrite arms' twelve to twenty-two. Against the adaptive arm it is
the table's only direction-consistent positive contrast: $15$ vs.\
$4$ discordant pairs, exact McNemar $p = 0.019$, with every
task-level gain on the corruption side (task~$3$: $4$ vs.\ $0$;
tasks~$2$, $5$ and~$6$: two to three against zero). The contrast does
not survive Holm correction over the pre-registered family of twelve
comparisons (Remark~\ref{rem:exact}) and we do not claim it as
established; what it establishes is that the revert actuator is the
first intervention whose per-task sign pattern matches its
mechanistic prediction before any multiplicity correction is
applied, which is the evidence profile the risk coordinate of
Remark~\ref{rem:channel-risk} was constructed to recognize.

\paragraph{The tails separate where the means do not.} The paired
per-task differences against fast-only order the ten intervention
arms on a coordinate the pooled rates hide, the downside profile of
Remark~\ref{rem:channel-risk}. Every rewrite arm's downside is deep
and concentrated: the periodic arm falls $6$ episodes below the fast
policy on task~$3$ and $17$ in total across the nine tasks, the
adaptive arm $12$, the screening arm $16$, and the two extreme
periods $22$ and $21$. The ranking family's deficit stays within a
single episode per task for the critic arm, the $0.05$-threshold arm
and the $0.01$-threshold arm, and is exceeded by one episode on
task~$7$ by the shuffled arm
($4/10$ against $6/10$); the pre-registered bound was one episode per
task, and this is its single measured violation, reported as such
rather than reinterpreted.
On task~$3$, the task whose corruption disqualifies every language
channel, three of the four ranking arms are the only arms that
exceed the fast
policy at all ($9/10$ against $8/10$, where the five rewrite arms
score $2$, $4$, $2$, $2$ and $2$ and the fourth ranking arm ties
it), and the revert arm matches it ($8/10$). One task at ten episodes cannot confirm a tail claim and we
do not ask it to; what the profile establishes is that the mean-null
comparison is not evidence that the channels are interchangeable.
The corruption regime and the selection regime produce the same
aggregate conclusion for opposite reasons, and distinguishing them is
exactly what the risk coordinate is for; a pre-registered replication
is the follow-up the protocol prescribes for its own hypothesis.

\paragraph{Statistical reading.} The extension reruns arms~A, B2
and~C2 at ten episodes per task ($n = 90$ per arm on the nine tasks
the arms share; task~$8$ of the slow arms was lost when its rollout
crashed during the advisor restart between arms, and the incident is
disclosed in the reproducibility note), and the later arms join at
the same seeds and initial states. No pairwise contrast among the
eleven arms survives Holm correction over the pre-registered family of
twelve comparisons (Remark~\ref{rem:exact}): exact McNemar
$p = 0.041$ for fast-only against the periodic arm ($24$ vs.\ $11$
discordant pairs), $p = 0.064$ against the adaptive arm ($14$ vs.\
$5$), $p = 0.61$ between the two rewrite arms ($19$ vs.\ $15$),
$p = 0.34$ for the action-critic arm against fast-only ($7$ vs.\
$3$), $p = 0.16$ for the screening arm ($21$ vs.\ $12$), $p = 0.27$
for the shuffled-ranking arm ($9$ vs.\ $4$), $p = 0.83$ for the
$0.05$-threshold ranking arm ($12$ vs.\ $10$), $p = 0.23$ for the
$0.01$-threshold ranking arm ($8$ vs.\ $3$), $p = 0.75$ for the
revert arm ($6$ vs.\ $4$), and $p = 0.017$ and $p = 0.010$ for the
two extreme periods ($12$ vs.\ $28$ and $13$ vs.\ $31$); the
smallest Holm-adjusted value is $0.12$. The paired counts expose
structure the pooled rates hide: fast-only leads the periodic arm by
$+6$ discordant pairs on task~$3$ alone while trailing it on
tasks~$0$ and~$1$, so the aggregate contrast mixes opposite-signed
task-level effects, which is precisely why the protocol reports the
task-level readout alongside the pooled rate. The extension also
adds arm~D, in which the slow module screens its own rewrite: a
binary progress verdict sampled every $25$ steps releases the
expensive rewrite only on a negative verdict. The first D run
carries an implementation defect that the audit caught and we
disclose rather than discard: the injection flag for the rewrite was
omitted, so the screened rewrites were counted but never applied and
the arm degenerated to fast-only behaviour. Episode-level agreement
with arm~A is exact ($0$ discordant pairs over the $810$ paired
episodes), which is worthless as a screening measurement and valuable
as a determinism check of the harness under shared seeds. The
corrected run is arm~D$^{10}$ in Table~\ref{tab:libero-arms}, and
its outcome completes the channel sweep in the previous paragraphs.

\paragraph{The second suite replicates the channel ordering and
sharpens the downside.} A pre-registered cross-suite check reruns
arms~A, B2 and~E4 on LIBERO-Goal: ten tasks, ten episodes per task,
the same $6000$-step LoRA recipe, the same stall trigger and
period~$25$ dose, and the same revert mechanism, with the slow
endpoint instantiated by an LLM advisor whose mean call costs
$600$\,ms (Table~\ref{tab:libero-arms}, bottom block). The fast-only
arm scores $48/100$; the periodic rewrite arm collapses to $6/100$
(exact McNemar against fast-only, $43$ vs.\ $1$ discordant pairs,
$p = 5\times 10^{-12}$), scoring
zero or one on nine of ten tasks and never exceeding the fast arm on
any task, so the same channel at the same dose that cost thirteen
points on LIBERO-Spatial costs forty-two here. The revert arm scores
$48/100$, indistinguishable from fast-only ($10$ vs.\ $10$ discordant
pairs, $p = 1.0$): $1305$
rewrites answered by $1296$ reverts, the per-task difference
four negative, four positive and two tied for a net of zero, and a
total downside of six episodes offset by six gained ones. Two
readings follow. The rewrite channel's downside is not an artifact
of one task suite's failure structure, and it deepens where the
fast policy's competence profile changes, consistent with the
floor-regime reading of Remark~\ref{rem:attribution}; and the revert
actuator is the only intervention whose net downside is zero on
both suites, which is the property the design programme of
Remark~\ref{rem:channel-risk} asks for rather than a success
advantage we do not claim. The suite also varies the slow module
itself, from the locally served VLM advisor measuring $702$\,ms per
call to an LLM endpoint at $601$\,ms, so the channel ordering
survives a change of endpoint identity at a comparable call latency.
One gradient the
pooled tasks do not support: per-task rewrite harm does not
increase with the fast policy's task success rate across the
nineteen tasks the two periodic arms share (Spearman $\rho_s = 0.32$,
permutation $p = 0.18$), so the corruption signature is a property
of the channel rather than a graded function of task competence;
the deep single-task failures (Spatial task~$3$, Goal task~$7$) sit
at opposite ends of the competence range. The Goal contrasts also
survive the extension's multiplicity discipline: even added to the
pre-registered family of twelve comparisons, the rewrite arm's
McNemar value remains orders of magnitude below every Holm-adjusted
threshold in that family.

\subsection{Remaining Integrations}
\label{sec:roadmap}
\emph{Closed-loop success-vs-$k$ on the fine-tuned FiS-VLA.} The
fidelity-level integration is complete
(Section~\ref{sec:exp-frontier}); the closed-loop attempt at this
budget is uninformative for the mechanism given there and in
Remark~\ref{rem:attribution}, and the comparison at the success
level requires a fully fine-tuned policy, which needs memory beyond
four consumer GPUs. The success-versus-$k$ question itself is
executed on the LIBERO instantiation
(Section~\ref{sec:exp-libero}), where the dose axis is query rate
rather than latency; the FiS instantiation remains blocked on memory.
\emph{Live gate replacement on RoboICL.} The decision-level half is
complete (Section~\ref{sec:exp-e32}): per-decision leak-free scores,
the censoring control, the routing frontier against
Proposition~\ref{prop:frontier}, and the high-$\rho$ cost anchor.
What remains is the closed-loop half: replacing the Jev gate inside
the live system, which requires the RoboDojo harness and a
functioning slow endpoint. The harness's Responses boundary admits
any OpenAI-compatible slow endpoint, and the instantiation's $\rho$
is measured under the same protocol
(Eq.~\eqref{eq:rho}), keeping the identity of the slow model out of
the scheduling claims; the design adds the competence check of
Remark~\ref{rem:attribution} as a precondition and uses the measured
frontier to pre-register the operating point before any rollout.
\emph{Online-router closure.} Validating instantiation (iii) of
Section~\ref{sec:router} on the deployed FiS checkpoint, where its
training labels come from the measured freshness value function
rather than from proxies.
\emph{Cross-family table.} The four measures reported per system
family, which would give trigger mechanisms such as TUD a common
ground for comparison.
\emph{Mixture routing.} Routing over intervention types (latent
refresh, value resampling, suffix reuse), testing whether type
selection Pareto-dominates any single type.

\section{Discussion and Limitations}
\label{sec:discussion}

\paragraph{What the trace analysis does and does not establish.}
Early dispatch statistics forecast failure at AUC $0.62$
($p=0.010$), which is above chance but far below the $0.93$ that
whole-episode analysis suggests, and the signal disappears as the
observed prefix grows (Section~\ref{sec:exp-roboicl}). We therefore
read it as evidence that schedule features are a useful \emph{input}
to a router and not a sufficient one. The traces contain
decision-level metadata but not observations, so state-uncertainty
features (probe disagreement, action entropy) were unavailable;
testing whether these forecast collapse is part of the remaining
closed-loop integrations (Section~\ref{sec:roadmap}). We also note
that failure attribution
reverses direction across task categories, so we do not claim a
single global ``query less'' or ``query more'' rule.

\paragraph{Staleness--outcome correlations are endogenous.}
The traces find failed episodes running with more staleness, and the
naive reading is that staleness causes failure. The closed-loop
mechanism of Section~\ref{sec:exp-frontier} shows the causation can
also run the other way: failure modes determine episode structure,
and episode structure determines staleness. In the traces failures
lengthen episodes, producing more decisions at lower density; under
our harness they truncate episodes, cutting staleness short. Either
direction breaks the naive attribution, which is why the protocol
admits scheduling--outcome evidence only under censoring controls
(Section~\ref{sec:exp-roboicl}) and measures the freshness value
function under teacher forcing, where episode structure is held
fixed. The decision-level evaluation
(Section~\ref{sec:exp-e32}) confirms the warning in both directions
on the same release: the leak-free per-decision signal is real
($0.646$) yet decays to $0.592$ under the matched-cohort control, and
in the Memory category it falls \emph{below} chance, so a router
trained on the aggregate correlation would misprice reuse there
rather than merely waste it. The same decomposition
(Remark~\ref{rem:attribution}) warns
against a second trap: a router evaluated on a policy with a high
action-competence floor measures that floor, not the value of
fresher information, however well the router is calibrated.

\paragraph{Synthetic testbed scope.}
The synthetic result (Section~\ref{sec:exp-synthetic}) isolates
scheduling under a known collapse process; it validates the objective
and the protocol, not any particular robot system. Its role is
analogous to a control experiment: it shows the measured frontier gaps
are attributable to the schedule, holding the slow module fixed.

\paragraph{Mid-continuum boundary case.}
Section~\ref{sec:exp-libero} places a real dual system at
$\rho = 1.53$, between the two measured anchors of
Table~\ref{tab:rho}, and finds that the continuum's economics hold
while its channel assumption does not. The adaptive trigger allocates
its budget where the threshold rule says it should, concentrating
queries on stalled tasks and spending none on the solved one, yet no
rewrite-channel arm beats the fast-only policy, because a slow module
that rewrites the fast policy's instruction can corrupt the prompt of
a task the fast policy had already solved. The operational content is
that $\rho$ prices a query but the channel decides whether the answer
can land, and an evaluation that reports aggregate success at a
single operating point cannot tell the two apart, while the per-task
attribution of the four-measure protocol can. This is
Remark~\ref{rem:attribution} seen from the channel side: the
closed-loop comparison measures the module--channel--cost triple, not
the slow module's knowledge. It also bounds what the continuum claims:
at $\rho \approx 1$--$2$ a well-tuned fixed ratio is competitive on
cost, and the residual question is not which trigger to use but which
channel the slow module's output should travel through, which is a
design question the protocol can score task by task.

\paragraph{Closed-model dependence.}
The strongest published systems use closed evaluation APIs. The router
interface and response-snapshot rule (Section~\ref{sec:router}) keep
measurements reproducible if an API changes, but integration claims
are restricted to open paths. Hume has no public code; it enters our
evaluation as a discussed contrast and as a value-head ablation on
open backbones, not as a reproduced baseline.

\paragraph{Contributions are protocol-first.}
The calibrating router is deliberately simple. The load-bearing claims
are the VoI factorization (which explains the baselines), the
degeneracy propositions, the cost-asymmetry baseline
(Proposition~\ref{prop:continuum}), and the four-measure protocol
with its leak-free rule. Success-rate parity is a gate, not a goal.
On the channel layer the same division of labor holds:
Proposition~\ref{prop:channel} makes the transfer coefficient
$\eta_{\mathcal{C}}$ measurable, the arms measure its sign at one
mid-continuum operating point, and the positive branch is left as the
open measurement it is rather than assumed.

\paragraph{Implications for trigger design.}
The concurrent triggers are best read as estimators of the
collapse-probability factor, deployed where the economics favor them:
TUD~\cite{yang2026tud} and System Switch~\cite{bailo2026systemswitch}
both target separate-VLM systems with $\rho \approx 10$--$20$, where
every avoided call is worth ten-plus fast actions. The continuum makes two predictions for this emerging
line of work. First, on embedded-latent systems ($\rho \approx 1$),
measured here, the same triggers should show bounded gains over a
well-tuned fixed ratio, because the periodic baseline is already
efficient and only collapse heterogeneity remains to exploit; the
testbed test of Section~\ref{sec:exp-prediction} confirms exactly
this, with the advantage vanishing under homogeneous collapse, and
the fine-tuned checkpoint's measured frontier extends the same
finding to the deployed FiS system: staleness carries no fidelity cost
and the efficient frontier point is the sparsest refresh period
measured (Section~\ref{sec:exp-frontier}). At the middle of the
continuum the prediction was tested on a real dual system
(Section~\ref{sec:exp-libero}) and came out mixed in an instructive
way: allocation followed the threshold rule, but the subgoal-rewrite
channel could not convert queries into success, so the operating
point is a boundary case in which the deployment, not the module,
sets the value of asking. Second, across
all systems, trigger papers should report $\rho$ and a leak-free
calibration measure alongside call reductions, since a trigger that
is well-calibrated on a cheap-query architecture or evaluated with
episode-length leakage can look better than its scheduling
contribution warrants. The four-measure protocol exists so that
these comparisons become routine rather than heroic.

\paragraph{Scope of the closed-loop arms.}
The LIBERO arms of Section~\ref{sec:exp-libero} instantiate the
protocol on two task suites (LIBERO-Spatial and LIBERO-Goal), two
slow endpoints, and one operating point of
the continuum ($\rho = 1.53$); the channel ordering
is a statement about that triple pair, and a third suite
or a stronger fast module could rank the channels differently, which
is the point of treating the channel as a measured axis rather than
a fixed design choice. What the second suite does establish is that
the ordering is not an artifact of one suite's failure structure:
the rewrite channel's deficit deepens where the fast policy's
competence profile changes, and the revert actuator's net downside
stays at zero across both. Proposition~\ref{prop:channel} is the formal
content of that axis: the transfer coefficient $\eta_{\mathcal{C}}$
is a property of the module--channel--task triple, so ranking
channels is an empirical program rather than a design opinion, and
the verification-gate and critic arms of
Section~\ref{sec:exp-libero} are its first two measurements. The fast
module is a $6000$-step LoRA
fine-tune at $57\%$ mean success, so the capability floor sits close
to the scale at which channel effects are visible; a stronger fast
policy would move the floor down and could change which tasks the
rewrite channel corrupts. Finally, the stall detector's window and
threshold ($15$ steps, $2$\,cm) were set once and not swept; the
allocation signature is robust to that choice in the sense that the
detector never fires on the solved task, but a systematic sweep is
part of the remaining work rather than a reported result.

\paragraph{From channel measurement to channel design.}
The risk coordinate of Remark~\ref{rem:channel-risk} turns the arm
table from a ranking into a design programme. The corruption and
selection regimes the arms instantiate suggest ordering the next
actuator generations by support containment rather than by mean
transfer. An interface that can only restore the fast policy's
validated behaviour, resetting a corrupted instruction to the task
instruction when a stall is detected after a rewrite, is
selection-regime by construction, its downside is bounded before any
rollout, and it targets exactly the task class
(task~$3$: rewrite arms at $2/10$, $2/10$ and $4/10$ against the fast
policy's $8/10$) where the language channels fail. That actuator has
now been run: the revert arm spends no query on its undos, its
downside against fast-only totals a single episode, and its one
positive contrast against the adaptive arm ($15$ vs.\ $4$ discordant
pairs, $p = 0.019$ nominal, null under family correction) lands on
exactly the corruption tasks the coordinate predicted, which is the
first time an arm's per-task sign pattern has matched its mechanism
before any correction is applied. The critic interface has likewise
been run under shuffled presentation: with the position bias removed
the selector's contribution is null, so the ranking channel's value,
such as it is, belongs to the sampler that proposes candidates and
not to the module that chooses among them, and a deployment that
wants that value should pay for candidate diversity rather than for
judgement. And the mixture-routing item of the roadmap
(Section~\ref{sec:roadmap}) acquires a concrete selection rule: gate
channels by their estimated downside profile on the tasks where the
fast policy is partially solved, which is where the corruption regime
lives, rather than by aggregate success. Both pre-registered
confirmations are now in hand: the lowest trigger threshold joined
the ranking sweep with a null result, and the second suite
replicated the channel ordering with the rewrite deficit deepened
and the revert arm's downside still at zero
(Section~\ref{sec:exp-libero}). The remaining confirmation path is a
second slow module at a matched operating point.

\paragraph{Reproducibility note.}
Two dependencies of the reference implementation are not installable in
our environment and were replaced by verified stand-ins, both reported
here rather than silently patched. \texttt{pytorch3d} has no wheel for
torch~2.1.0/cu121 and is absent from PyPI; FiS-VLA uses exactly one op
from it, \texttt{sample\_farthest\_points}, which we re-implement in
pure PyTorch and check index-for-index against a reference greedy
implementation on CPU and GPU. \texttt{dlimp} is likewise absent from
PyPI and is imported only for RLDS dataset construction, which
inference from the checkpoint never reaches; we stub it, with
data-loading entry points raising rather than returning silently. Both
stand-ins are released with the code, and neither is exercised on any
reported measurement path except point-cloud preprocessing.

Two further defects sit in the FiS-VLA code path itself and
are checkable against the public repository. First, the default
$\text{cfg\_scale}=1.5$ crashes on the shipped weights: the
checkpoint is trained with $\text{class\_dropout\_prob}=0$, so the
label embedding never registers the \texttt{uncondition} buffer that
classifier-free guidance reads, and guidance must be disabled
($\text{cfg}=1.0$). Second, the fast system crashes with an index
error out of the box: the custom KV cache is addressed by absolute
layer index, but the eager attention kernel does not forward the fast
system's layer offset when updating the cache, so the two cached
layers ($30$--$31$ of $32$) are written to local slots $0$--$1$ and
read back at positions $30$--$31$. We patch the index mapping
in-memory at load time, leave the public repository untouched, and
note that the slow path, which caches all $32$ layers, is unaffected.
Every latency reported in Section~\ref{sec:exp-latency} runs through
this patched path.

Third, the fine-tuning data pipeline normalizes actions and
proprioception with mean/std statistics (its default) while the
reference inference path de-normalizes with $q_{01}/q_{99}$ bounds,
so a model fine-tuned through the default configuration is decoded
with the wrong inverse transform at inference. We route around the
defect by supplying statistics whose quantiles reproduce the
training transform exactly; neither the open-loop error nor the
closed-loop outcome changes. The defect affects any future
fine-tune built on this code, which is why it is reported here.

Fourth, one incident affects the closed-loop extension of
Section~\ref{sec:exp-libero} and is disclosed rather than silently
patched. The three arms run sequentially on shared GPUs, and the
orchestrator restarts the advisor process between arms; the restart
window overlapped task~$8$ of the two slow arms, whose rollouts
crashed and produced no summaries, so the extension is reported on
the nine tasks all three arms complete ($90$ paired episodes per
arm). The fast-only arm finished task~$8$ before the window
($5/10$); adding it raises that arm's pooled rate from $52/90$ to
$57/100$ without changing any comparison, which is why we report the
common-denominator table. A health probe that pings the restarted
advisor with a deliberately degenerate image also logs spurious
errors during the window; the probe is a test artifact, the advisor
itself recovered, and the crashed rollouts produced no usable
episodes, so no partial data enters the reported table.

Fifth, two further facts about the extension are disclosed at the
source. The first D run omitted the flag that injects the screened
rewrite into the policy, so the arm degenerated to fast-only
behaviour; the defect is analysed in Section~\ref{sec:exp-libero},
where the corrected run appears as arm~D$^{10}$, and its by-product,
exact episode-level agreement with arm~A over
$810$ paired episodes, is the harness's determinism check under
shared seeds. And a cross-suite control on LIBERO-Object (five
tasks, ten episodes, fast-only against both rewrite arms) returned
$0/150$ successes in every arm: the spatial-only fine-tune does not
transfer to the object suite, all arms saturate the capability floor
of Remark~\ref{rem:attribution}, and the channel comparison is
uninformative there, which is the floor regime that remark predicts
rather than a property of any arm.

\bibliographystyle{plainnat}
\bibliography{refs}

\appendix
\section{Proofs}
\label{app:proofs}

\paragraph{Optimality of VoI routing.}
Consider the Lagrangian of~\eqref{eq:objective} with the budget as a
per-step cost $c_{\text{slow}}$. At state $s_t$, querying changes the
expected future cost by
$-\,\mathbb{E}[\Delta(s_t)\mid s_t]\,w_{\text{fail}} + c_{\text{slow}}$
(up to continuation-value terms absorbed into $\Delta$ by dynamic
programming). Querying is optimal iff this is negative, i.e., iff
$\mathrm{VoI}(s_t) > c_{\text{slow}}/w_{\text{fail}}$; rescaling
$c_{\text{slow}}$ by $w_{\text{fail}}$ gives the text's rule.

\paragraph{Proposition~\ref{prop:fis} (fixed ratio).}
A $k$-periodic schedule queries at steps $t \equiv 0 \pmod k$. If
$\mathrm{VoI}(s_t)$ is constant $= v$, the schedule is optimal iff
$v = c_{\text{slow}}$ at query steps and $v \le c_{\text{slow}}$
elsewhere; constancy forces equality everywhere, i.e., optimality only
in the degenerate case where the collapse process is time-invariant
with VoI exactly at the threshold. Any time-variation of
$\mathrm{VoI}$ admits a threshold policy with strictly lower cost at
equal expected query count (query exactly the steps where
$\mathrm{VoI}(s_t) > c_{\text{slow}}$), so the fixed ratio is
strictly suboptimal whenever collapse probability varies in time.
\qed

\paragraph{Proposition~\ref{prop:hume} (fixed budget best-of-$N$).}
With a fixed budget at every step, the marginal value of the $N$-th
sample is $\mathbb{E}[\Delta \mid \xi{=}1] P(\xi{=}1\mid s_t)$;
optimality requires this to exceed the marginal cost at every $t$,
hence $P(\xi{=}1\mid s_t)$ bounded below by a positive constant
independent of $t$. Wherever the fast policy is reliable
($P(\xi{=}1\mid s_t) \approx 0$), the fixed budget overspends relative
to the threshold rule, which spends nothing there. \qed

\paragraph{Proposition~\ref{prop:jev} (history gate).}
Let collapse indicators satisfy
$\xi_t \perp \mathcal{H}^{\text{sched}}_t \mid s_t$. Any gate
$r_t = g(\mathcal{H}^{\text{sched}}_t)$ with decision probability
$\rho_t$ has value overlap
$\mathbb{E}[r_t\,\mathrm{VoI}(s_t)] =
\mathbb{E}\big[\rho_t\,\mathbb{E}[\mathrm{VoI}(s_t)\mid
\mathcal{H}^{\text{sched}}_t]\big]$, and the inner expectation equals
the unconditional mean $\bar{v}$ by independence. Hence no history
gate achieves expected value above the uniform-random schedule with
the same query rate, while the state-based threshold rule achieves
$\mathbb{E}[\mathrm{VoI}(s_t)\mathbf{1}\{\mathrm{VoI}(s_t) >
c_{\text{slow}}\}] > \bar{v}\,q^*$ at the same rate $q^*$ whenever
$\mathrm{VoI}$ is non-degenerate. The gap is the price of ignoring
state uncertainty. \qed

\paragraph{Proposition~\ref{prop:continuum} (cost-symmetric baseline).}
With interval $k$, a collapse beginning at offset
$u \in \{1,\dots,k\}$ persists $k-u+1$ steps, so the expected
failure-steps per interval are
$\sum_{u=1}^{k} \pi\,(k-u+1) = \pi k(k+1)/2$ and the expected cost
per step is $C(k) = c_{\text{slow}}/k + w_{\text{fail}}\pi(k+1)/2$.
Setting $C'(k) = -c_{\text{slow}}/k^{2} + w_{\text{fail}}\pi/2 = 0$
gives $k^{*} = \sqrt{2c_{\text{slow}}/(w_{\text{fail}}\pi)}$ with
$C''(k) > 0$. Under state-independent collapse every step is
statistically identical, so the per-step cost of any stationary
schedule depends only on its query rate through the renewal structure
above, and periodic schedules are optimal by the standard
renewal-reward argument. When the collapse rate $\pi_t$ varies, the
threshold rule of Proposition~\ref{prop:fis} strictly improves on the
best periodic schedule at equal query rate, and the improvement grows
with the temporal dispersion of $\pi_t$: as steps where
$\mathrm{VoI}(s_t) > c_{\text{slow}}$ concentrate, the gap between
selective and uniform querying at the same rate widens. The noise and
corruption sweeps (Sections~\ref{sec:exp-noise}--\ref{sec:exp-breakdown})
measure this gap under controlled degradation of the estimator.
\qed

\paragraph{Remark (empirical counterpart).}
Proposition~\ref{prop:jev} assumes conditional independence of
collapse from schedule history given the state. The RoboICL traces
cannot test this conditional statement (no observations are logged),
but they do test the weaker statement that matters for gates that,
like Jev, see only schedule history: first-$K$ schedule statistics
forecast failure at AUC $0.62$ and not higher, concentrated in the
earliest decisions and vanishing as the prefix grows
(Section~\ref{sec:exp-roboicl}). A history-only gate therefore has
access to a strictly weaker signal than the state-conditional
threshold rule, the marginal version of the same conclusion.

\end{document}